# Integrating Factual and Normative Industrial Knowledge via Constraint-Aware Graph Attention for Process Plan Recommendation

Yuntong Chen[a], Yingqi Li[a], Yingying Xiao[b], Ziang Wang[a], Zewei Liu[b], Jiahao Liu[b], Xitian Tian[a], Lijiang Huang[a]*

[a] School of Mechanical Engineering, Northwestern Polytechnical University, Xi'an, Shaanxi, 710072, China

[b] State Key Laboratory of Complex Product Intelligent Manufacturing System Technology, Beijing Institute of Electronic System Engineering, Beijing, 100854, China

**Corresponding authors: Lijiang Huang[a]***

## Abstract

**Integrating heterogeneous industrial knowledge — factual relationships among entities and normative constraints governing decisions — remains a core challenge in industrial information systems. Machining process planning exemplifies this challenge, as engineers must select appropriate operations by synthesizing material properties, feature characteristics, and quality requirements — yet existing recommendation methods rely on similarity retrieval or classification without a unified ranking-learning objective or standardized evaluation protocol, hindering knowledge transfer between the engineering and recommendation systems communities. We propose PCA-GAT, a framework that casts machining process plan recommendation as a Knowledge Graph (KG)-enhanced Collaborative Filtering (CF) problem. The CF framework provides a principled ranking-learning objective — Bayesian Personalized Ranking (BPR) — and standardized evaluation protocol (Recall@K, NDCG@K); the KG serves as the primary semantic bridge, since collaborative signals alone prove insufficient at engineering-scale interaction volumes. Four types of domain constraints — material compatibility, precision requirements, feature applicability, and operation sequencing — are injected as continuous-valued attention biases during graph propagation. Type-specific learnable weights automatically discover constraint importance, while an adaptive gating mechanism modulates constraint influence based on each edge's local context. On a real-world aerospace engineering dataset (115 parts, 507 plans), PCA-GAT achieves Recall@1 = 0.9087 and demonstrates exceptional cold-start resilience, with Recall@1 degradation roughly half that of the strongest baseline under severe data sparsity. Ablation confirms that KG enrichment is indispensable, constraints provide additional value, and adaptive gating is essential — raw constraint injection without gating hurts performance. The learned weights identify material–operation compatibility as the dominant factor, consistent with domain expertise. Experiments on three public benchmarks confirm zero degradation when constraints are absent, demonstrating that the constraint-aware architecture generalizes as an engineering knowledge integration paradigm beyond manufacturing. To our knowledge, this is the first study to establish standardized recommendation evaluation protocols in the engineering process recommendation domain, systematically benchmarking seven methods spanning three categories. This cross-category evaluation reveals that the primary bottleneck is knowledge representation rather than algorithmic capacity.**

# 1. Introduction

Knowledge-driven decision support in engineering domains requires the integration of heterogeneous knowledge types: factual relationships among entities and normative constraints governing decisions. With the rapid development of knowledge engineering technologies, knowledge graphs have become the mainstream approach for organizing such knowledge [1-4], particularly in manufacturing — a domain where process planning exemplifies this knowledge integration challenge, serving as the core link connecting design and production [5,6]. It is essentially a knowledge-intensive decision-making activity where engineers must synthesize a vast amount of information (including material properties, machining equipment capabilities, geometric feature characteristics, and quality requirements) to make appropriate decisions regarding machining operations, cutting tools, and process parameters [7-9]. However, the efficient reuse of process knowledge faces significant challenges in current industrial practice: knowledge is scattered across technical documents, enterprise databases, and engineers' experiences, suffering from weak correlations and structural fragmentation [10-12]. Traditional keyword-based retrieval methods are inefficient and often fail to return relevant knowledge when queries do not perfectly match stored keywords [7]. Furthermore, static knowledge bases cannot effectively respond to dynamically changing design requirements [13]. These issues have prompted researchers to adopt the Knowledge Graph (KG) as the mainstream technical approach for organizing and representing manufacturing knowledge [1,2]. Recent studies have further explored the possibility of integrating large language models (LLMs) with KGs to enhance knowledge management efficiency [14,15].

In recent years, KG-based process planning research has made significant progress in both construction and application dimensions. In terms of construction, representative methods such as LLM-driven [8], dynamically updatable [13], and bidirectional knowledge extraction [16] have emerged. Application paradigms cover multi-layer KG reasoning [9], Graph Neural Network (GNN)-driven decision support [5], similarity-matching scheme selection [17], and process design for complex electronic products [18] (detailed in Section 2.1). However, these methods exhibit high heterogeneity in their technical routes—ranging from similarity-based retrieval matching [8,16], classification or sequence generation modeling [19,20], and reinforcement learning path reasoning on KGs [21], to the recent context-aware knowledge recommendation proposed by Wu et al. [22]—yet consistently lack a common modeling paradigm and evaluation benchmark.

The direct consequence of this heterogeneity is that meaningful quantitative comparisons between different methods are impossible. Because various works adopt different task definitions (retrieval, classification, path reasoning) and evaluation methods (classification accuracy, path accuracy, qualitative case studies), many fundamental questions remain unanswered. For example, is pure collaborative filtering effective in manufacturing scenarios? What is the actual gap between existing process recommendation methods and the mature technologies of the recommendation system community? In fact, the recommendation system community has developed mature Collaborative Filtering (CF) frameworks over the past decade to address such issues: providing unified optimization objectives through pairwise ranking losses (such as BPR (Bayesian Personalized Ranking)), and enabling fair comparisons between methods through standardized evaluation protocols (train/test splits, Recall@K, NDCG@K).

Based on this, this paper argues that incorporating machining process plan recommendation into a KG-enhanced CF framework serves two purposes: it introduces standardized modeling and evaluation paradigms that enable cross-community methodology transfer, and more fundamentally, it provides a principled engineering informatics framework in which factual knowledge (KG topology), normative knowledge (domain constraints), and empirical knowledge (historical interactions) can be explicitly represented and jointly utilized — achieving a workflow transition from passive 'human seeking knowledge' to active 'knowledge seeking human'. It should be pointed out that, due to the extremely limited interaction scale in manufacturing scenarios, what truly plays a critical role is the semantic bridge provided by the KG, while the core value of the CF framework lies in providing unified optimization objectives and evaluation benchmarks, rather than the collaborative signals themselves.

However, directly migrating recommendation system technologies to manufacturing process planning is not a simple domain adaptation, but rather faces three challenges unique to manufacturing scenarios:

(1) Interaction data is inherently sparse, with available historical records far fewer than in consumer recommendation scenarios, making it difficult for pure collaborative filtering methods to learn meaningful representations.

(2) Decision-making is driven by hard process constraints—such as material compatibility, precision requirements, feature mapping, and operation sequences—which carry continuous strengths and polarities, and the purely data-driven attention mechanisms of existing KG-enhanced recommendation methods cannot distinguish between factual knowledge and normative knowledge.

(3) Engineers require traceable recommendation explanations, while the attention paths of existing methods reflect statistical associations rather than domain rules.

To address the above three challenges, this paper proposes PCA-GAT (Process Constraint-Aware Graph Attention Network), a recommendation framework that explicitly injects domain-specific process constraints into the graph attention propagation process. Its core idea is to treat constraint knowledge as an additional signal channel independent of the KG topology, imposing continuous-valued biases in the attention computation, enabling the model to enhance attention on constraint-compatible edges and suppress attention on constraint-violating edges. Centered around this idea, the framework's design unfolds along four stages: "data foundation → knowledge encoding → propagation mechanism → training optimization". First, to compensate for the lack of interaction data, we merge part-plan interaction data with the process KG into a unified Collaborative Knowledge Graph (CKG), establishing multi-hop connections between parts and plans through shared semantic attributes such as materials, features, and operations. Based on this, we formalize four types of domain constraints (material-operation, precision-operation, feature-operation, and operation sequence) into a constraint matrix carrying continuous strength values in [-1, +1], and expand coverage through bidirectional expansion and 1-hop KG propagation. Subsequently, during the graph attention propagation process, type-specific learnable weights automatically discover the relative importance of various constraints, and an adaptive gating mechanism dynamically adjusts the constraint injection strength based on the local context of edge endpoints, avoiding the imposition of uniform constraint biases on all edges. Notably, this constraint-aware design naturally derives traceable recommendation explanations: type-specific weights reflect the relative importance of various constraints, and constraint activation paths reveal the specific engineering rules driving the recommendations. Finally, the model is trained end-to-end combining BPR ranking loss, TransE KG embedding loss, and constraint alignment regularization loss. Unlike KGAT [27], which learns attention weights purely from data, PCA-GAT enhances data-driven attention with explicit constraint bias channels, preserving the continuous strength, polarity, and context-dependent importance of domain rules.

The main contributions of this paper are summarized as follows:

**(1) Methodological Contribution:** We propose a constraint-aware graph attention mechanism that processes factual and normative engineering knowledge through independent channels. Four types of process constraints are injected as continuous-valued attention biases during graph propagation via type-specific learnable weights and adaptive gating networks. This mechanism enables the model to distinguish between descriptive knowledge ("is-related-to") and prescriptive knowledge ("should-use") and dynamically adjust constraint impact strength based on the local context of each edge — a capability absent from existing KG-enhanced recommendation methods. The learnable type weights further provide constraint-based traceable explanations for recommendation results. The design is domain-agnostic: any engineering domain with codified rules can instantiate the constraint matrix without modifying the model architecture.

**(2) Paradigm and Evaluation Contribution:** We are the first to incorporate machining process plan recommendation into a KG-enhanced collaborative filtering framework and establish standardized recommendation evaluation protocols (train/test splits, Recall@K, NDCG@K) in the engineering process recommendation domain. Under this protocol, we systematically benchmark seven methods spanning three categories, providing a methodological bridge between domain-specific engineering and the recommendation systems community. This protocol reveals a finding invisible to prior evaluation approaches: the primary bottleneck in engineering process recommendation is knowledge representation rather than algorithmic capacity — pure CF methods fail entirely despite interaction density 16× higher than consumer benchmarks.

**(3) Empirical Contribution:** Systematic experiments on a real-world aerospace engineering dataset and three public benchmarks yield three findings with implications beyond manufacturing. First, PCA-GAT achieves optimal accuracy under full data and demonstrates exceptional cold-start resilience, with Recall@1 degradation roughly half that of the strongest baseline under severe data sparsity — confirming that domain constraints function as data-independent inductive priors whose value scales inversely with data availability. Second, ablation studies reveal that adaptive gating is essential: raw constraint injection without gating degrades performance below the constraint-free baseline, establishing that context-adaptive modulation is a prerequisite for beneficial constraint integration. Third, experiments on three public benchmarks confirm zero performance penalty in constraint-absent scenarios, demonstrating cross-domain applicability. The model requires only sub-MB memory and minute-level training time, and a prototype system compliant with the ISA-95 standard has been validated with real-world enterprise data.

# 2. Related Work

This section reviews three related research directions: manufacturing domain knowledge graphs and process recommendations (Section 2.1), GNN-based KG-enhanced recommendations (Section 2.2), and domain knowledge integration in recommendation systems (Section 2.3), building upon which we distill the research gaps driving this proposed method (Section 2.4).

## 2.1 Knowledge Graph for Manufacturing and Process Recommendation

Knowledge graphs have become a powerful technical paradigm for organizing and representing manufacturing knowledge. Wan et al. [1] comprehensively reviewed key research topics of KGs in smart manufacturing, emphasizing their capabilities in semantic analysis, reasoning, and decision support. Xiao et al. [2] systematically reviewed the current state of KG-based manufacturing process planning technologies, covering the complete technical chain from knowledge representation and extraction to KG construction and process generation. Recent review works have further outlined research progress in this field from the perspective of factory physics [30] and the dimensions of knowledge acquisition, representation, and application [31].

At a specific technical level, recent studies have made significant progress in both KG construction and KG application. In terms of construction, Xu et al. [8] introduced an LLM-driven approach that reduced construction time and costs, while Zhang et al. [13] proposed dynamically updatable KGs for CAPP (Computer-Aided Process Planning) to respond to changing production demands. Ma et al. [6] combined KGs with deep learning to build a machining process knowledge base, and Xiao et al. [16] constructed assembly process KGs by bidirectionally extracting knowledge from historical process documents. Recently, Zhou et al. [14] proposed an LLM-based intelligent management method for manufacturing KGs, covering automated construction, refinement, and natural language question answering. Yan et al. [32] improved the integration method between LLMs and KGs, enhancing the automatic construction quality of machining process knowledge bases.

In terms of applications, the mKGMPP framework proposed by Huang et al. [9] organizes process knowledge across multiple dimensions and enables interactive reasoning. Su et al. [5] implemented KG-driven manufacturing decision support based on GNNs. Liu et al. [3] proposed a multi-hierarchical graph convolutional network for industrial KG embeddings. Zhu et al. [10] applied KG planning to remanufacturing processes, while Xue et al. [18] expanded KG technology to the intelligent design of debugging processes for complex electronic products. Zhang et al. [33] integrated similarity computation with KGs for the automatic recognition of machining and inspection features. Regarding process plan selection, Wang et al. [17] and Li et al. [34] conducted plan recommendations based on improved cosine similarity matching and constructed process KGs, respectively. Xiao et al. [35] proposed KGESM, a similarity matching model based on KG embeddings, which utilizes semantic similarities in the embedding space to generate assembly processes; however, it essentially remains a content-based retrieval paradigm.

However, as shown in Table 1, existing process recommendation methods have two systematic limitations. First, methodologically, they are confined to retrieval and classification paradigms: current works either match historical process routes via similarity [17,34] or model the problem as a classification [19] or sequence reasoning [21] task, none of which utilize the collaborative filtering signal of "similar parts selecting similar plans". Second, evaluation lacks standardization, a limitation that has somewhat restricted methodological advancement in this field. Specifically, among the 6 representative works in Table 1: 2 adopted qualitative case validations without any quantitative baseline comparisons [9,21]; 2 solely compared KGE methods (e.g., TransE vs. TransR) without involving collaborative filtering or retrieval baselines [5,34]; 1 used classification accuracy instead of ranking metrics [19]; and only 1 adopted recommendation evaluation metrics (F1 and NDCG), but its user scale was limited to 13 users [22]. Currently, no existing work reports Recall@K and NDCG@K under train/test splits while simultaneously benchmarking across multiple categories against representative methods in the recommendation systems field (such as KGAT, KGIN, and LightGCN). This state of evaluation makes a fundamental question difficult to answer: faced with the mature technologies already present in the recommendation system community, what is the actual gap for existing manufacturing process recommendation methods?

Table 1. Comparison of evaluation approaches in manufacturing process recommendation.

| Work | Task Type | Dataset Scale | Evaluation Metric | Evaluation Protocol | Recommendation Baselines |
|---|---|---|---|---|---|
| Li et al. 2025 [34] | KG similarity retrieval | 1 target part | HR/MRR (300 trials) | No train/test split; single target part | None (similarity only) |
| Huang et al. 2025 [9] | Multi-layer KG reasoning | 1 case study | Qualitative | Case study; no baselines | None |
| Zhou et al. 2025[21] | KG path reasoning (RL) | Case study | Path accuracy 0.81 | Case study; KGE-only baselines | TransE/TransR only |
| Z. Wang et al. 2024 [19] | Feature-level classification | 559 parts | Classification 93.3% | Train/test; CNN/GCN variants only | CNN/GCN variants |
| Su et al. 2025 [5] | KG link prediction | KG reasoning | MRR, Hits@K | KGE methods only; no CF baselines | KGE methods only |
| Wu et al. 2026 [22] | Context-aware retrieval | 700 docs, 13 users | F1-score, NDCG | 13 users; content-based baselines | Content-based, UCF |
| **Our work** | **KG-enhanced CF (plan-level ranking)** | **115 parts, 507 plans** | **Recall@K, NDCG@K** | Standardized train/test split; 7 methods across 3 categories | **7 methods, 3 categories** |

## 2.2 KG Enhanced Recommendation Based on GNN

Graph Neural Networks (GNNs) have become the dominant paradigm in recommendation systems due to their ability to model high-order interactions through neighborhood aggregation [25]. A recent survey [41] systematically reviewed the application of GNNs in recommendations from the perspective of multi-type scenarios. In purely collaborative filtering settings, LightGCN [26] demonstrated that simplified graph convolutions retaining only neighborhood aggregation can achieve excellent performance; however, pure CF methods rely entirely on interaction data and fail when data is insufficient.

To resolve data sparsity, KG-enhanced GNN methods integrate knowledge graphs as auxiliary information sources into recommendations. KGAT [27] is a pioneering work in this direction: it builds a Collaborative Knowledge Graph (CKG) by merging user-item interaction graphs with KGs and applies learnable attention weights to propagate embeddings across the unified graph. KGIN [28] further models user intents as combinations of KG relations, providing finer-grained preference decomposition. Subsequent works introduced contrastive learning to enhance representation quality: MCCLK [36] aligns knowledge and interaction graphs via cross-view contrastive learning, while KGCL [37] reduces data noise through knowledge-guided graph contrastive learning. Diffusion models [38] and LLMs have also been explored for KG-enhanced recommendations. Notably, KG4RecEval [39] discovered from an evaluation perspective that many models, including KGAT, do not utilize KG knowledge efficiently—after randomly deleting KG facts, the recommendation accuracy of most models did not consistently decline—which further inspired the need for more effective KG information utilization methods.

Although these methods have achieved notable success in consumer recommendation scenarios, applying them to the manufacturing domain presents two core limitations. First, attention learning relies on a sufficient absolute volume of interactions. KGAT's attention mechanism can indeed learn the importance differences of various edges from data, but this learning requires the support of substantial interaction signals. In industrial manufacturing scenarios, available historical interaction records typically measure in the hundreds, far below the hundreds of thousands or even millions found in consumer recommendation datasets; this discrepancy in magnitude poses fundamental difficulties for purely data-driven attention learning. Second, and more fundamentally, purely data-driven attention cannot distinguish between factual knowledge and normative constraint knowledge. Edges in a KG represent factual relationships (e.g., "45 steel IS-A carbon steel"), whereas process constraints represent normative knowledge (e.g., "45 steel SHOULD-USE turning"), with the latter carrying continuous strength and polarity. These two types of semantically distinct knowledge should not be processed by the same data-driven mechanism. Even with sufficient interaction data, KGAT's attention cannot "discover" continuous strength values of material compatibility from interaction signals because these values originate from engineering domain knowledge rather than user behavior patterns.

## 2.3 Domain Knowledge Integration in Recommendation

There are three main paradigms for integrating domain knowledge into recommendation systems [23,40], categorized by the stage and granularity of constraint participation in learning: post-processing filtering applies global constraints after inference, regularization enforces global constraints during training, and graph enhancement applies edge-level constraints during training. However, all three lack context-adaptive constraint adjustment.

"Post-processing filtering" first generates a recommendation list and then filters out unqualified items using domain rules. This method is simple and does not affect model training, but constraints cannot influence representation learning [36]—when the model learns embeddings without constraint guidance, plans violating constraints might be placed near the target part in the embedding space. While post-processing filtering can remove them, it cannot correct the already distorted embedding geometry.

"Regularization and auxiliary loss" encodes domain knowledge as additional loss terms to constrain the embedding space. This incorporates domain knowledge into the learning process but acts at a global embedding level, lacking the fine-grained edge-level control required for manufacturing constraints. Uta et al. [23] proposed the concept of constraint-based recommender systems, but focused on constraint satisfaction of user preferences rather than constraint injection within graph propagation.

"Graph structure enhancement" encodes constraints as additional KG edges or relation types, naturally blending them into graph propagation. However, constraint edges are far fewer than existing KG edges, and their influence is gradually diluted through multi-layer GNN propagation [37]. Furthermore, binary edge representations (presence/absence) cannot capture continuous constraint strengths and polarities.

The shared core deficiency across these three paradigms is the lack of context-adaptive constraint adjustment: the same constraint rule exerts uniform influence across all edges, ignoring specific part-plan contexts. In manufacturing, however, the relevance of a constraint is highly context-dependent—for instance, "high-precision surfaces require grinding" should be strictly enforced for aerospace parts, but may not be mandatory for general-purpose parts.

## 2.4 Summary of Research Gaps

Based on the above review, we identify three progressive research gaps that collectively drive the design of the PCA-GAT framework.

**Gap 1 (Dual absence of paradigm and evaluation):** Manufacturing process recommendation has neither been incorporated into collaborative filtering frameworks nor established standardized recommendation evaluation protocols. As shown in Table 1 and Section 2.1, no prior work has modeled part-plan matching as a KG-enhanced collaborative filtering problem. More importantly, the evaluation methods of existing works—case studies, KGE method comparisons, or classification accuracies—cannot be meaningfully compared with mature methods in the recommendation system community. This paper fills this gap through two complementary contributions: (a) establishing a CF framework in the manufacturing scenario for the first time; and (b) introducing standardized recommendation evaluation protocols (train/test splits, Recall@K, NDCG@K) and systematically comparing against seven baseline methods across three categories, providing a methodological bridge between the manufacturing domain and the recommendation systems community.

**Gap 2 (Absence of mechanism):** Existing KG-enhanced recommendation methods cannot distinguish between factual knowledge and normative constraint knowledge. As mentioned in Section 2.2, methods like KGAT treat all KG edges equally, learning attention weights purely from data. As discussed in Section 2.3, encoding constraints as KG edges leads to signal dilution, while post-processing filtering and regularization lack edge-level control. The continuous strengths, polarities, and type-specific importance of normative process constraints cannot be effectively captured in existing frameworks. This paper fills this gap by proposing a constraint-aware attention bias mechanism with type-specific learnable weights.

**Gap 3 (Absence of optimization):** Existing constraint integration methods lack context-adaptive constraint adjustments. As analyzed in Section 2.3, all three paradigms apply uniform constraint influence to all edges, ignoring the specific part-plan context. The lack of contextual adjustment not only limits the effectiveness of constraints but could even introduce harmful biases. This paper fills this gap by designing an adaptive gating network that dynamically calculates constraint injection strength based on the embeddings of the edge endpoints.

Importantly, these gaps are not unique to manufacturing: any engineering domain where decisions follow codified rules with continuous strengths — such as clinical treatment planning, structural engineering design, or regulatory compliance checking — faces the same challenge of integrating factual and normative knowledge within graph-based decision systems.

These three gaps form a progressive chain of "Paradigm → Mechanism → Optimization", collectively defining the design space of PCA-GAT. The next section details the methodology, starting with problem modeling and the construction of the collaborative knowledge graph.

# 3. Methodology

This section presents the proposed Process Constraint-Aware Graph Attention Network (PCA-GAT) for machining process plan recommendation. We first formulate the problem and provide a framework overview, then detail each component: collaborative knowledge graph construction, process constraint modeling, constraint-aware graph attention mechanism, and model training.

## 3.1 Notations and Framework Overview

### 3.1.1 Problem Formulation

Let $G_{KG}$=(E,R,T) denote the machining process knowledge graph, where E is the set of entities (including materials, geometric features, machining operations, equipment, etc.), R is the set of relation types, and $T=\{(h,r,t)|h,t\in E,r\in R\}$ is the set of factual triplets.

Let $P=\{p_1,p_2,\ldots,p_M\}$ denote the set of parts and $S=\{s_1,s_2,\ldots,s_N\}$ denote the set of candidate process plans. The historical part-plan adoption records are represented as a binary interaction matrix $\mathbf{Y}\in\{0,1\}^{M\times N}$, where $y_{ij}=1$ indicates that plan $s_j$ has been adopted for part $p_i$.

In addition, we define four types of domain-specific process constraints:

$$C=\{C_{mat},C_{prec},C_{feat},C_{seq}\}$$

where $C_{mat}$ denotes material-operation constraints, $C_{prec}$ denotes precision-operation constraints, $C_{feat}$ denotes feature-operation constraints, and $C_{seq}$ denotes operation sequence constraints. Each constraint carries a strength value in [-1,1], where positive values indicate compatibility and negative values indicate incompatibility.

**Problem Definition.** Given a target part $p$ with its associated attributes in $G_{KG}$, the constraint set$C$, and the historical interactions $\mathbf{Y}$, the goal is to learn a scoring function $f(p,s)$ that predicts the relevance of each candidate process plan $s\in S$ to the target part, such that the top-ranked plans are most suitable for the part's manufacturing requirements.

Table 2. Summary of key notations.

| Notation | Description |
|---|---|
| $G_{KG}$ | Process knowledge graph |
| $G_{CKG}$ | Collaborative knowledge graph |
| E,R,$T$ | Entity set, relation set, triplet set |
| $P$,$S$ | Part set, process plan set |
| Y | Part–plan interaction matrix |
| $e_i^{(l)}$ | Embedding of entity $i$ at layer $l$ |
| $r_{ij}$ | Relation embedding for edge $(i,j)$ |
| $d$ | Embedding dimension |
| $L$ | Number of GAT propagation layers |
| $C_c$ | Constraint set of type $c$ |
| $C_{ij}^c$ | Constraint score of type cc c for edge $(i,j)$ |
| $\lambda_g$ | Bounded global constraint strength |

| Notation | Description |
|---|---|
| $\lambda_c$ | Bounded weight of the c-th constraint type |
| $g_{ij}^{(l)}$ | Adaptive gate value for edge $(i,j)$ at layer $l$ |
| $\alpha_{ij}^{(l)}$ | Attention weight for edge $(i,j)$ at layer $l$ |
| $L_{BPR}$ | Bayesian Personalized Ranking loss |
| $L_{KG}$ | Knowledge graph embedding loss |
| $L_{align}$ | Constraint alignment regularization loss |

3.1.2 Framework Overview

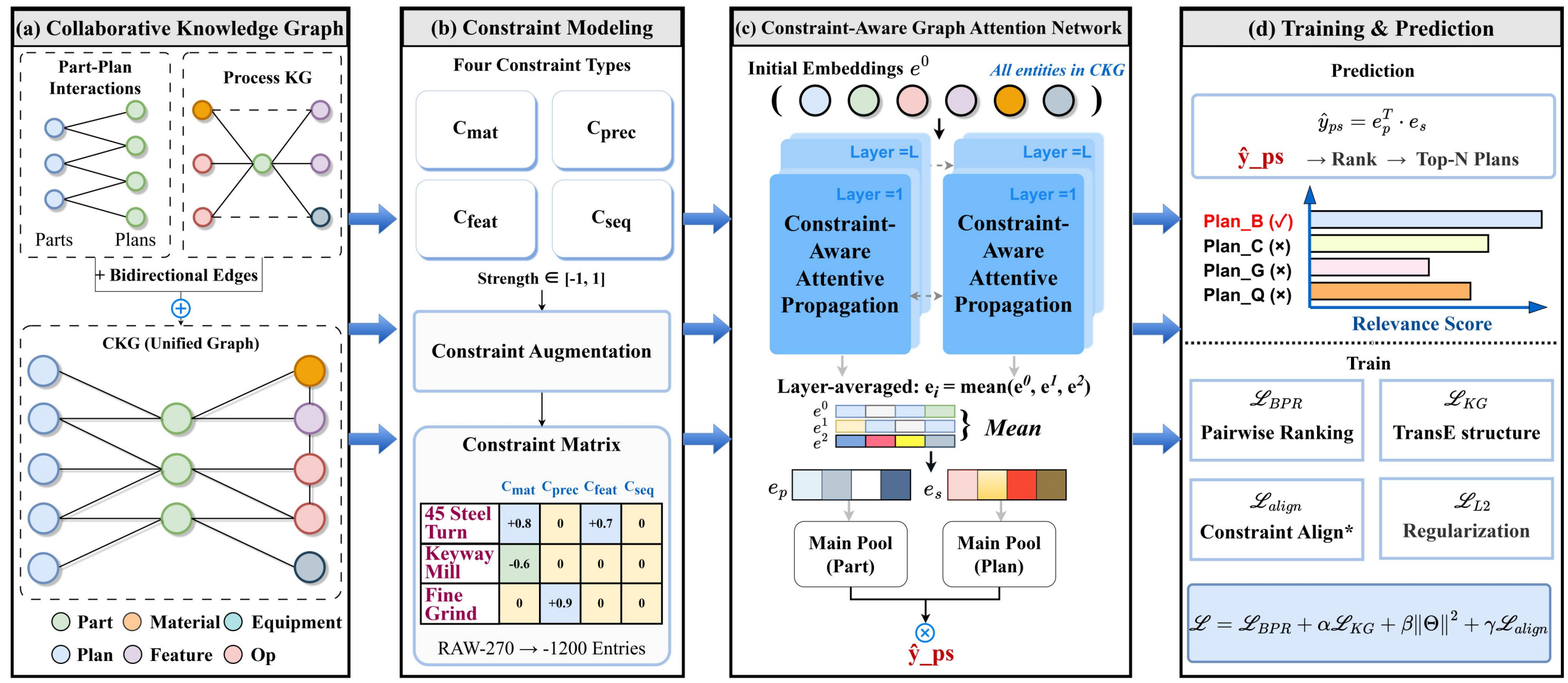

Fig. 1. Overall framework of PCA-GAT

As illustrated in Fig. 1, the proposed PCA-GAT framework consists of four main components:

(1) Collaborative Knowledge Graph Construction (Section 3.2). We construct a unified collaborative knowledge graph (CKG) by merging the part-plan interaction graph with the process knowledge graph, where parts and plans are connected to rich semantic entities such as materials, features, operations, and equipment.

(2) Process Constraint Modeling (Section 3.3). Four types of domain-specific process constraints are formally encoded into a constraint matrix. For each edge in the CKG, a constraint score $C_{ij}^{c} \in [-1,1]$ is assigned based on the applicable constraint rules. Constraint augmentation via bidirectional expansion and 1-hop KG propagation is applied to improve coverage.

(3) Constraint-Aware Graph Attention Network (Section 3.4). The core of our approach: a modified graph attention mechanism that injects process constraints as attention biases during message passing. An adaptive gating mechanism dynamically controls the constraint injection strength based on the local context of each edge. The detailed architecture of the propagation layer and the constraint bias injection mechanism are presented in Fig. 2.

(4) Model Training and Prediction (Section 3.5). The model is trained end-to-end with BPR loss, TransE KG embedding loss, and constraint alignment regularization. During inference, the learned embeddings are used to compute relevance scores and generate ranked recommendation lists.

## 3.2 Collaborative Knowledge Graph Construction

### 3.2.1 Process Knowledge Graph

The process knowledge graph $G_{KG}$ captures the structured relationships among manufacturing entities involved in process planning. The entity types include: parts (workpieces characterized by material grade, blank type, and geometric complexity), process plans (complete manufacturing solutions specifying operation sequences, equipment, and parameters), materials (e.g., 45 steel, 7075 aluminum alloy), geometric features (e.g., outer cylindrical surfaces, inner holes, keyways, end faces), machining operations (e.g., turning, milling, grinding, drilling, heat treatment), equipment (e.g., lathes, milling machines, CNC machining centers), and tooling (cutting tools, fixtures, measuring instruments).

The relation types capture semantic associations such as "has_material", "contains_feature", "uses_operation", "performed_on", and "requires_tool". The KG is populated from enterprise manufacturing databases and domain handbooks, with all triplets verified by process engineers.

### 3.2.2 Collaborative Knowledge Graph (CKG)

Following the collaborative knowledge graph paradigm [27], we construct the CKG by merging the part-plan interaction graph with the process knowledge graph:

$$G_{CKG}=\{(p,interact,s)|y_{ps}=1\}\cup\{(\mathrm{h},r,t)|(\mathrm{h},r,t)\in T\}$$

To enable bidirectional information flow, each KG triplet $(\mathrm{h},r,t)$ is augmented with its inverse $(t,r^{-1},\mathrm{h})$, and each interaction edge is likewise made bidirectional. The relation embedding vocabulary therefore covers $2|\mathrm{R}|+1$ types: original relations, inverse relations, and the interaction relation.

This semantic enrichment is particularly critical for our machining scenario, where the absolute volume of interaction data is extremely limited (371 training interactions, compared to 846,434 in Amazon-Book), despite a relatively high interaction density (0.64% vs. 0.04%).Without the KG, collaborative filtering methods have insufficient signal to learn meaningful representations, as we demonstrate in Section 4.3.

This characteristic is shared by most engineering decision-support scenarios, where historical records are inherently limited compared to consumer domains, making KG-based semantic enrichment a general prerequisite rather than a domain-specific design choice.

## 3.3 Process Constraint Modeling

While the CKG captures factual knowledge about manufacturing entities and their relationships, it does not encode the normative knowledge that governs process planning decisions. A CKG edge between "45 steel" and "turning" tells us they are related, but not whether turning is appropriate for 45 steel under specific precision requirements, or whether turning should precede or follow heat treatment. This normative knowledge—the "should" rather than the "is"—is captured by process constraints.

### 3.3.1 Four Types of Process Constraints

We define four types of constraints that collectively govern machining process decisions:

Type 1: **Material-Operation Constraints** $C_{mat}$. These encode the compatibility between material types and machining operations. For example, "45 steel is suitable for turning" (strength +0.8), or "cast iron is unsuitable for high-speed milling without coolant" (strength −0.6). The strength ranges from strong recommendation (+1.0) to strong prohibition (−1.0).

Type 2: **Precision-Operation Constraints** $C_{prec}$. These encode the relationship between surface quality or tolerance requirements and the operations needed to achieve them. For instance, "surface roughness Ra ≤ 0.8 μm requires grinding" (strength +0.9).

Type 3: **Feature-Operation Constraints** $C_{feat}$. These encode the mapping between geometric features and applicable operations. For example, "inner hole features require boring or drilling" (strength +0.9), or "keyway features require slotting or broaching" (strength +0.85).

Type 4: **Operation Sequence Constraints** $C_{seq}$. These encode the required ordering between operations. For instance, "quenching must precede grinding" (strength +1.0, directional).

Each constraint is represented as a tuple $(e_i,e_j,c_{ij})$, where $e_i,e_j \in E$ are entity pairs and $c_{ij} \in [-1,1]$ is the constraint strength. Positive values indicate compatibility or required association; negative values indicate incompatibility or prohibited association.

**Constraint annotation protocol.** Two senior process engineers (each with over 15 years of experience) independently annotated all constraint entries using a 5-level discrete scale (+1.0, +0.5, 0.0, −0.5, −1.0), with intermediate values at 0.1 intervals permitted. A two-round procedure was followed: (i) independent annotation, and (ii) reconciliation for entries differing by more than 0.3. Inter-annotator agreement reached ICC = 0.87 (95% CI: [0.83, 0.91]) before reconciliation. The final constraint set contains 68 material-operation, 55 precision-operation, 72 feature-operation, and 75 operation-sequence entries, totaling 270 raw entries. The annotated data and guidelines will be released with the dataset.

### 3.3.2 Constraint Matrix Construction

For each edge $(i,j)$ in the CKG and each constraint type $c \in \{mat, prec, feat, seq\}$, we assign a type-specific constraint score:

$$C_{ij}^{c} = \begin{cases} s_{ij}^{c} & if\,(e_i,e_j) \in C_c \\ 0 & otherwise \end{cases}$$

Each edge thus carries a four-dimensional constraint profile $\mathbf{C}_{ij} = (C_{ij}^{mat}, C_{ij}^{prec}, C_{ij}^{feat}, C_{ij}^{seq})$. The aggregate constraint bias is:

$$b_{ij} = \sum_{c=1}^{4} \lambda_c \cdot C_{ij}^{c}$$

where $\lambda_c \in (0,1)$ is the bounded learnable weight for constraint type $c$ (Section 3.4.3). Edges with $\boldsymbol{C}_{ij}^{c} = 0$ (no applicable constraint of any type) receive $b_{ij} = 0$, relying entirely on data-driven attention.

### 3.3.3 Constraint Augmentation

Manually defined constraints have limited coverage (~270 entries vs. >10,000 CKG edges). We apply two augmentation strategies:

**Bidirectional Expansion.** For each non-directional constraint ($c \in \{mat, prec, feat\}$), we add the reverse $(e_j,e_i,c_{ij})$ where it does not already exist. Operation sequence constraints are excluded due to their inherent directionality.

**1-Hop KG Propagation.** For each constraint ($e_i$,$e_j$,$c_{ij}$), we propagate to 1-hop KG neighbors with decay factor $\beta$=0.5. Propagated constraints with absolute strength below threshold $\tau$=0.1 are filtered out. This expands coverage from approximately 270 to approximately 1,200 entries while maintaining semantic validity.

## 3.4 Constraint-Aware Graph Attention Network

This section presents the core technical contribution: a graph attention mechanism that integrates process constraints through attention biases with adaptive gating. As illustrated in Fig. 2, the propagation layer consists of two tightly coupled components: (a) the constraint-aware attentive propagation layer, which computes attention-weighted neighborhood aggregation with residual connections and layer normalization; and (b) the constraint bias injection mechanism, where type-specific learnable weights $\lambda_c$ aggregate four constraint channels into a scalar bias $b_{ij}$, and a context-dependent gate $g_{ij}$ dynamically modulates the injection strength before the bias is added to the data-driven attention logit $z_{ij}$.

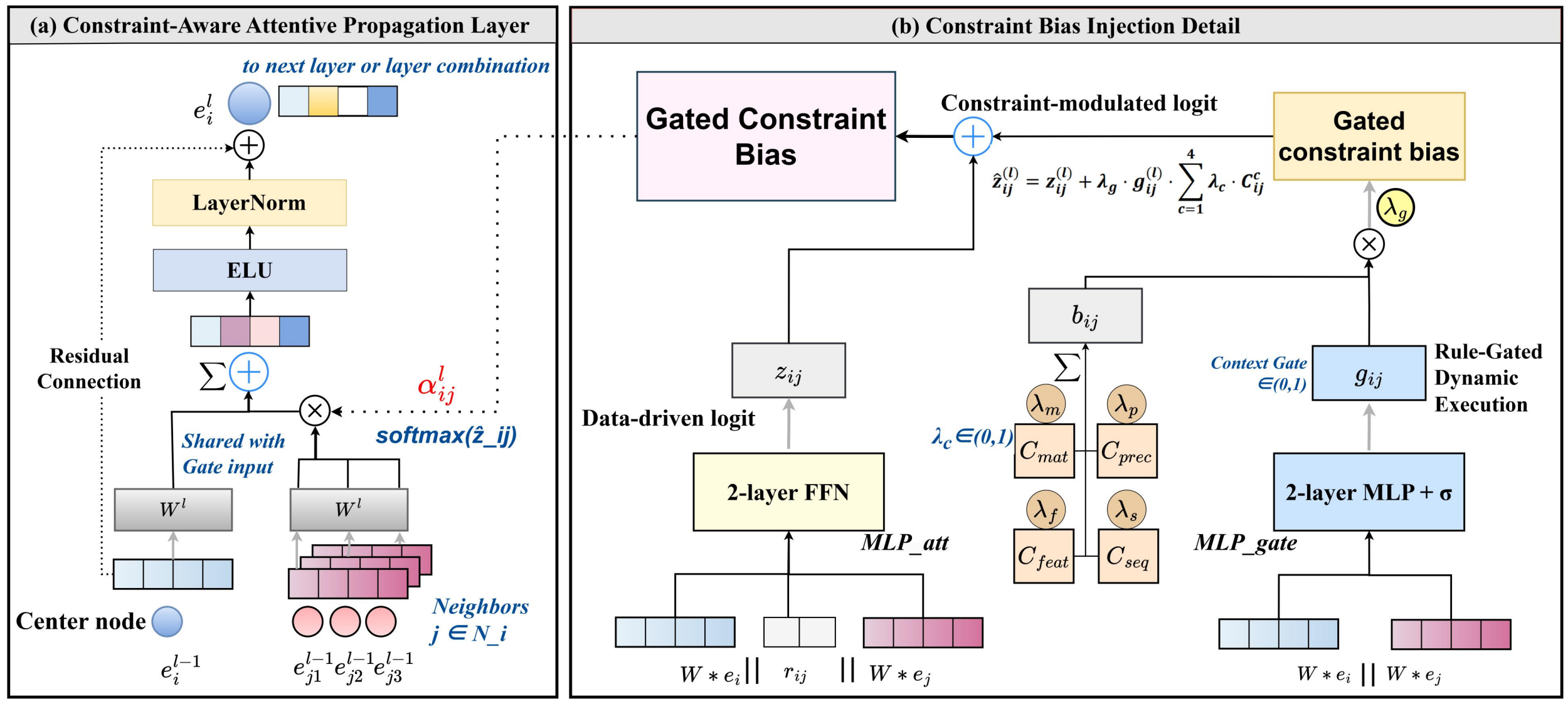

Fig. 2. Detail of the constraint-aware attentive propagation layer. (a) Propagation layer architecture with attention-weighted aggregation, residual connection, and layer normalization. (b) Constraint bias injection mechanism: four constraint types are aggregated via learnable weights $\lambda_c$ into a scalar bias, modulated by an adaptive context gate $g_{ij}$, and added to the data-driven attention logit.

### 3.4.1 Embedding Layer

Each entity $e_i$∈E is initialized with a learnable embedding vector $\mathbf{e}_i^{(0)}$∈R$^d$. Each relation $r$∈R is associated with a learnable embedding $\mathbf{r}$∈R$^d$. All embeddings are initialized using Xavier uniform initialization.

### 3.4.2 Relation-Aware Attention Mechanism

Unlike standard GAT, which computes attention coefficients solely from node features, our attention mechanism additionally incorporates relation-type information. For each edge $(i,j)$ with relation type $r_{ij}$ in the CKG, the attention logit at layer $l$ is computed as:

$$z_{ij}^{(l)}=MLP_{att}^{(l)}\left(\left[\mathbf{W}^{(l)}\mathbf{e}_i^{(l\text{-}1)}\|\mathbf{W}^{(l)}\mathbf{e}_j^{(l\text{-}1)}\|\mathbf{r}_{ij}\right]\right) \quad \text{Eq. (1)}$$

where $\mathbf{W}^{(l)}\in\mathrm{R}^{d'\times d}$ is the feature transformation matrix, $\|$ denotes concatenation, $\mathbf{r}_{ij}\in\mathrm{R}^{d}$ is the embedding of the relation type on edge $(i,j)$, and $MLP_{att}^{(l)}$ is a two-layer feed-forward network:

$$MLP_{att}^{(l)}(\mathbf{x})=\mathbf{w}_2^{(l)\intercal}\cdot LeakyReLU\left(\mathbf{W}_1^{(l)}\mathbf{x}+\mathbf{b}_1^{(l)}\right) \quad \text{Eq. (2)}$$

with $\mathbf{W}_1^{(l)}\in\mathrm{R}^{d'\times(2d'+d)}$, $\mathbf{w}_2^{(l)}\in\mathrm{R}^{d'}$, and $\mathbf{b}_1^{(l)}\in\mathrm{R}^{d'}$.

The inclusion of the relation embedding $\mathbf{r}_{ij}$ enables the attention mechanism to differentiate edges based on their semantic relationship types, which is important in the heterogeneous CKG where edges carry diverse semantics.

### 3.4.3 Constraint-Aware Attention with Type-Specific Learnable Weights

To inject process constraint knowledge, we modify the attention logit by adding a gated constraint bias term:

$$\hat{z}_{ij}^{(l)}=z_{ij}^{(l)}+\lambda_g\cdot g_{ij}^{(l)}\cdot\sum_{c=1}^{4}\lambda_c\cdot C_{ij}^{c} \quad \text{Eq. (3)}$$

Where $z_{ij}^{(l)}$ is the data-driven attention logit from Eq. (1); $C_{ij}^{c}\in[-1,1]$ is the constraint score; $g_{ij}^{(l)}\in[0,1]$ is the adaptive gate value (Section 3.4.4); $\lambda_c$ and $\lambda_g$ are bounded learnable parameters (detailed below).

The attention weight is then obtained by normalizing the modulated logits over each node's neighborhood:

$$\alpha_{ij}^{(l)}=\frac{\exp\left(\hat{z}_{ij}^{(l)}\right)}{\sum_{k\in N_i}\exp\left(\hat{z}_{ik}^{(l)}\right)} \quad \text{Eq. (4)}$$

Bounded learnable constraint weights. We use the sigmoid function to ensure stable training and interpretable values:

$$\lambda_c=\sigma(\hat{\lambda}_c)\in(0,1),\quad c=1,\ldots,4 \quad \text{Eq. (5)}$$

$$\lambda_g=\sigma(\hat{\lambda}_g)\cdot S\in(0,S) \quad \text{Eq. (6)}$$

where $\hat{\lambda}_c$ and $\hat{\lambda}_g$ are the raw (unbounded) learnable parameters, and $S = 2$ is the maximum global constraint strength. This bounding prevents the constraint alignment loss from trivially minimizing itself, provides directly interpretable importance values, and offers implicit regularization through sigmoid saturation.

### 3.4.4 Adaptive Gating Mechanism

A critical insight from our experiments is that injecting raw constraints without modulation can harm performance (Section 4.3, w/o Gate variant). The same constraint rule may have different relevance in different contexts — e.g., "high-precision surfaces require grinding" should be strongly enforced for aerospace parts but may be less critical for general-purpose parts.

The gate is computed from the transformed embeddings of the two endpoint nodes:

$$g_{ij}^{(l)} = \sigma\left(MLP_{gate}^{(l)}\left(\left[\mathbf{W}^{(l)}\mathbf{e}_i^{(l\text{-}1)} \| \mathbf{W}^{(l)}\mathbf{e}_j^{(l\text{-}1)}\right]\right)\right) \quad \text{Eq. (7)}$$

where $MLP_{gate}^{(l)}$ is two-layer MLP with LeakyReLU activation:

$$MLP_{gate}^{(l)}(\mathbf{x}) = \mathbf{w}_{g2}^{(l)\top} \cdot LeakyReLU\left(\mathbf{W}_{g1}^{(l)}\mathbf{x} + \mathbf{b}_{g1}^{(l)}\right) + b_{g2}^{(l)} \quad \text{Eq. (8)}$$

The gate input uses the transformed embeddings $\mathbf{W}^{(l)}\mathbf{e}_i^{(l\text{-}1)}$ rather than the raw embeddings. Because the gate decision should be made in the same representational space where attention is computed. The gate value encodes the semantic context of the edge endpoints, learning to open (≈1.0) when a constraint is highly relevant and close (≈0.0) when a constraint is less relevant or potentially misleading.

### 3.4.5 Message Passing and Layer Aggregation

With the constraint-aware attention weights, the message passing at layer $l$ incorporates a residual connection:

$$\tilde{\mathbf{e}}_i^{(l)} = ELU\left(\sum_{j \in N_i} \alpha_{ij}^{(l)} \cdot \mathbf{W}^{(l)}\mathbf{e}_j^{(l\text{-}1)} + \mathbf{W}^{(l)}\mathbf{e}_i^{(l\text{-}1)}\right) \quad \text{Eq. (9)}$$

The output is stabilized through layer normalization and a cross-layer residual connection:

$$\mathbf{e}_i^{(l)} = LayerNorm\left(\tilde{\mathbf{e}}_i^{(l)}\right) + \mathbf{e}_i^{(l\text{-}1)} \quad \text{Eq. (10)}$$

Dropout is applied to both attention weights and aggregated features during training. After $L$ layers, the final representation is obtained by averaging across all layers:

$$\mathbf{e}_i = \frac{1}{L+1} \sum_{l=0}^{L} \mathbf{e}_i^{(l)} \quad \text{Eq. (11)}$$

This layer combination strategy, following [26], retains information from different propagation depths. The 0-th layer captures the original entity features, the 1st layer captures direct neighbor information, and higher layers capture progressively more global structural patterns modulated by constraints.

## 3.5 Model Training and Prediction

### 3.5.1 Prediction Function

The relevance score between part $p$ and process plan $s$ is computed by the inner product of their final embeddings:

$$\hat{y}_{ps}=\mathbf{e}_p^{\intercal}\cdot\mathbf{e}_s \qquad \text{Eq. (12)}$$

### 3.5.2 BPR Recommendation Loss

We adopt the Bayesian Personalized Ranking (BPR) loss [42], which learns to rank positive (adopted) plans higher than negative (non-adopted) plans:

$$\mathrm{L}_{BPR}=-\sum_{(p,s^+,s^-)\in D}\ln\sigma\left(\hat{y}_{ps^+}-\hat{y}_{ps^-}\right) \qquad \text{Eq. (13)}$$

where $D=\{(p,s^+,s^-)|y_{ps^+}=1,y_{ps^-}=0\}$ is the set of training triplets, $s^+$ is a positive plan adopted by part $p$, $s^-$ is a randomly sampled negative plan, and $\sigma(\cdot)$ is the sigmoid function.

### 3.5.3 Knowledge Graph Embedding Loss

To enhance the quality of entity representations, we jointly train a KG embedding objective using the TransE model [44]. For each KG triplet (h,$r$,$t$) and a corrupted triplet (h,$r$,$t'$) obtained by replacing the tail entity with a random entity $t'$, the TransE loss is:

$$\mathrm{L}_{KG}=\sum_{(\mathrm{h},r,t)\in T}\sum_{(\mathrm{h},r,t')\in T'}\max\left(0,\|\mathbf{e}_\mathrm{h}+\mathbf{r}-\mathbf{e}_t\|_1-\|\mathbf{e}_\mathrm{h}+\mathbf{r}-\mathbf{e}_{t'}\|_1+m\right) \qquad \text{Eq. (14)}$$

where $\mathbf{r}_r$ is a dedicated relation embedding for TransE (separate from the relation embeddings used in the attention mechanism), and $m$ is the margin hyperparameter. |

### 3.5.4 Constraint Alignment Regularization Loss

To provide a direct gradient pathway for learning $\lambda_c$ and $\lambda_g$ (which participate in attention only indirectly through softmax), we introduce:

$$\mathrm{L}_{align}=-\lambda_g\cdot E_{(i,j)\sim C}\left[\sum_{c=1}^{4}\lambda_c\cdot\cos\left(\bar{\mathbf{e}}_i,\bar{\mathbf{e}}_j\right)\cdot C_{ij}^{c}\right] \qquad \text{Eq. (15)}$$

where $\bar{\mathbf{e}}$ denotes detached (stop-gradient) entity embeddings. Key design choices: (1) detached embeddings ensure $\mathrm{L}_{align}$ only updates $\lambda$ values without interfering with primary embedding learning; (2) the bounded $\lambda$ values directly multiply the loss, creating a self-regulating dynamic where the model increases $\lambda$ for constraint types where embedding alignment agrees with constraint signs; (3) mini-batch sampling (2,048 pairs per step) ensures computational efficiency.

### 3.5.5 Total Training Objective

The total training loss is a weighted combination of the four loss terms:

$$L = L_{BPR} + \alpha \cdot L_{KG} + \beta \cdot \|\Theta\|_2^2 + \gamma \cdot L_{align} \quad \text{Eq. (16)}$$

where $\Theta$ denotes all learnable parameters (embeddings, attention MLP weights, transformation matrices, gate parameters, and constraint weights), $\alpha$ is the KG loss weight, $\beta$ is the L2 regularization coefficient, and $\gamma$ is the alignment loss weight.

### 3.5.6 Training Strategy

We employ differentiated learning rates: constraint-related parameters ($\hat{\lambda}_c$, $\hat{\lambda}_g$, and gate MLP parameters) are optimized with a learning rate $\kappa$ times that of the main parameters ($\kappa$=10 by default). Gradient norm clipping (max norm 5.0) prevents gradient explosion. Early stopping monitors Recall@K on the validation set with a minimum of $T_{min}$ warmup epochs. The complete training procedure is presented in Algorithm 1.

### 3.5.7 Training Algorithm

**Algorithm 1: PCA-GAT Training**

1. **Input:**
2. CKG graph $G_{CKG}$, constraint sets $C=\{C_{mat},C_{prec},C_{feat},C_{seq}\}$, interaction data Y, embedding dim $d$, layers $L$, learning rates $\eta$, $\eta_c=\kappa\cdot\eta$, weights $\alpha$, $\beta$, $\gamma$
3. **Output:**
4. Trained model parameters $\Theta$
5. ---
6. **// Initialization**
7. Initialize embeddings $\{\mathbf{e}_i^{(0)}\}$, relation embeddings $\{\mathbf{r}_k\}$ with Xavier uniform
8. Initialize MLP parameters $\{\mathbf{W}^{(l)}, MLP_{att}^{(l)}, MLP_{gate}^{(l)}\}$ for $l$=1,…,$L$
9. Initialize raw constraint weights $\hat{\lambda}_g$, $\{\hat{\lambda}_c\}_{c=1}^4$
10. Construct expanded constraint matrix via bidirectional expansion and 1-hop KG propagation (Section 3.3.3)
11. **// Training**
12. for *epoch*=1 to $T_{\max}$ do
13. Sample BPR mini-batch $B=\{(p,s^+,s^-)\}$ from Y
14. Sample KG mini-batch $B_{kg}=\{(h,r,t,t')\}$ from $T$
15. // Compute bounded constraint weights
16. $\lambda_c \leftarrow \sigma(\hat{\lambda}_c)$ for $c$=1,…,4; $\lambda_g \leftarrow \sigma(\hat{\lambda}_g)\cdot S$
17. // GNN Propagation (L layers)
18. $H^{(0)} \leftarrow$[entity_emb]
19. for $l$=1 to $L$ do
20. for each edge $(i,j)$ with relation $r_{ij}$ in $G_{CKG}$ do

21. $z_{ij} \leftarrow MLP^{(l)}_{att}\left([\mathbf{W}^{(l)}\mathbf{e}_i^{(l\text{-}1)} \| \mathbf{W}^{(l)}\mathbf{e}_j^{(l\text{-}1)} \| \mathbf{r}_{ij}]\right)$ // Eq. (1)
22. $b_{ij} \leftarrow \sum_{c=1}^{4} \lambda_c \cdot C_{ij}^c$ $\quad$
23. $g_{ij} \leftarrow MLP^{(l)}_{gate}\left([\mathbf{W}^{(l)}\mathbf{e}_i^{(l\text{-}1)} \| \mathbf{W}^{(l)}\mathbf{e}_j^{(l\text{-}1)}]\right)$ // Eq. (7)
24. $\hat{z}_{ij} \leftarrow z_{ij} + \lambda_g \cdot g_{ij} \cdot b_{ij}$ // Eq. (3)
25. end for
26. $\alpha_{ij} \leftarrow softmax_{j \in N_i}(\hat{z}_{ij})$ // Eq. (4)
27. $\tilde{\mathbf{e}}_i^{(l)} \leftarrow ELU\left(\sum_{j \in N_i} \alpha_{ij} \cdot \mathbf{W}^{(l)}\mathbf{e}_j^{(l\text{-}1)} + \mathbf{W}^{(l)}\mathbf{e}_i^{(l\text{-}1)}\right)$ // Eq. (9)
28. $\mathbf{e}_i^{(l)} \leftarrow LayerNorm\left(\tilde{\mathbf{e}}_i^{(l)}\right) + \mathbf{e}_i^{(l\text{-}1)}$ // Eq. (10)
29. end for
30. $\mathbf{e}_i \leftarrow \frac{1}{L+1} \sum_{l=0}^{L} \mathbf{e}_i^{(l)}$ $\quad$ // Eq. (11)
31. // Compute losses
32. $L_{BPR} \leftarrow - \sum_{(p,s^+,s^-) \in B} \ln \sigma\left(\mathbf{e}_p^{\top}\mathbf{e}_{s^+} - \mathbf{e}_p^{\top}\mathbf{e}_{s^-}\right)$ // Eq. (13)
33. $L_{KG} \leftarrow \sum_{(h,r,t,t') \in B_{kg}} \max\left(0, \|\mathbf{e}_h + \mathbf{r}_r - \mathbf{e}_t\|_1 - \|\mathbf{e}_h + \mathbf{r}_r - \mathbf{e}_{t'}\|_1 + m\right)$ // Eq. (14)
34. $L_{align} \leftarrow -\lambda_g \cdot mean\left[\sum_c \lambda_c \cdot \cos(\bar{\mathbf{e}}_i, \bar{\mathbf{e}}_j) \cdot C_{ij}^c\right]$ over sampled pairs // Eq. (15)
35. $L \leftarrow L_{BPR} + \alpha \cdot L_{KG} + \beta \cdot \|\Theta\|_2^2 + \gamma \cdot L_{align}$ // Eq. (16)
36. // Parameter update with differentiated learning rates
37. $Clip \|\nabla\Theta\|$ to max norm 5.0
38. Update $\Theta \setminus \{\hat{\lambda}_g, \hat{\lambda}_c, \Theta_{gate}\}$ with learning rate $\eta$
39. Update $\{\hat{\lambda}_g, \hat{\lambda}_c, \Theta_{gate}\}$ with learning rate $\eta_c = \kappa \cdot \eta$
40. // Early stopping
41. if $epoch \geq T_{\min}$ and $epoch \bmod T_{eval} = 0$ then
42. Evaluate Recall@K on validation set
43. if no improvement for $P$ consecutive evaluations: break
44. end if
45. end for
46. // Inference
47. Propagate through $L$ layers (same as training forward pass, without dropout)
48. For target part $p$, rank all candidate plans by $\hat{y}_{ps} = \mathbf{e}_p^{\top} \cdot \mathbf{e}_s$
49. return top-$N$ plans

# 4. Experiments and Results

This section presents comprehensive experiments to evaluate the proposed PCA-GAT framework. We aim to answer the following research questions:

**RQ1 (Overall Performance):** How does PCA-GAT perform compared to state-of-the-art recommendation methods across different datasets?

**RQ2 (Ablation Study):** What is the contribution of each key component (KG, constraint attention, adaptive gating)?

**RQ3 (Constraint Analysis):** How do different constraint types, coverage, and strength affect recommendation performance?

**RQ4 (Parameter Sensitivity):** How sensitive is PCA-GAT to key hyperparameters?

**RQ5 (Explainability):** Can PCA-GAT provide constraint-driven explanations that align with domain expert reasoning?

## 4.1 Experimental Setup

### 4.1.1 Datasets

We evaluate PCA-GAT on three datasets: one domain-specific machining process dataset and 3 widely-used public recommendation benchmarks. The domain-specific dataset validates the core value of constraint-aware recommendation in manufacturing, while the public datasets demonstrate that PCA-GAT maintains competitive performance in general scenarios where constraints are absent.

**Machining Process Dataset.** We construct a real-world machining process dataset from an aerospace manufacturing enterprise, containing historical records of part-plan adoption decisions. Each part is represented as a node in the CKG, connected to its material, geometric features, and quality requirements through KG relations. Each process plan specifies a sequence of operations, equipment, and parameters, represented as a subgraph in the KG. Four types of process constraints are annotated by senior process engineers and augmented via the strategies described in Section 3.3.3.This scale is representative of real-world part families within a single manufacturing enterprise and, as shown in Table 1, constitutes one of the largest quantitative evaluations in manufacturing process recommendation to date.

**Public Benchmarks.** Amazon-Book, Yelp2018, and Last-fm [27] are widely-used KG-enhanced recommendation benchmarks. We follow the standard preprocessing and train/test splits of [27] for fair comparison.

Table 3. Dataset statistics.

| | Machining | Amazon-Book | Yelp2018 | Last-fm |
|---|---|---|---|---|
| Users (Parts) | 115 | 70679 | 45919 | 23,566 |
| Items (Plans) | 507 | 24915 | 45,538 | 48,123 |
| Interactions | 371 (train) + 136 (test) | 846434 | 1,185,068 | 3,023,657 |
| Interaction Density | 0.64% | 0.04% | 0.04% | 0.27% |
| KG Entities | 1049 | 113487 | 136499 | 58,266 |
| KG Relations | 10 | 39 | 42 | 9 |
| KG Triplets | 4639 | 2557746 | 1853704 | 464,567 |
| Constraints (raw) | ~270 | N/A | N/A | N/A |
| Constraints (augmented) | ~1,200 | N/A | N/A | N/A |

Note: Process constraints are specific to the machining domain and are not applicable to the public datasets. On public datasets, PCA-GAT operates with $C_{ij}^{c}=0$ for all edges, effectively reducing to a standard GAT baseline, which isolates the effect of the constraint mechanism.

### 4.1.2 Baseline Methods

We compare PCA-GAT against baselines spanning three categories to provide a comprehensive evaluation:

**Category 0: Manufacturing-Domain Baselines** (retrieval/embedding without collaborative filtering). These two baselines represent the predominant methodologies currently adopted in manufacturing process recommendation literature (see Table 1), neither of which leverages collaborative filtering signals.

1.CB-KG (Content-Based KG Retrieval): (Content-Based KG Retrieval): A non-parametric method that constructs attribute neighborhood profiles from the CKG and retrieves candidate plans by Jaccard similarity over KG attribute sets, representative of the similarity-based paradigm [17,34].

2.TransE-Rec (KG Embedding Retrieval): Entity embeddings learned via TransE [44] on the process KG, with plans ranked by cosine similarity (d=64, lr=0.001, margin=1.0, 500 epochs), representative of the KG embedding approach [5,10].

**Category 1: Collaborative Filtering (no KG).**

3.BPR-MF [42]: Bayesian Personalized Ranking with matrix factorization. A classic CF method that learns user/item embeddings from pairwise ranking loss without any side information.

4.LightGCN [26]: A simplified GNN for collaborative filtering that removes feature transformations and nonlinear activations, retaining only neighborhood aggregation. Represents state-of-the-art pure CF.

**Category 2: KG-Enhanced Recommendation.**

5.CKE [43]: Collaborative Knowledge base Embedding that jointly learns from CF signals and KG structural/textual/visual information through TransR embeddings.
6.KGAT [27]: Knowledge Graph Attention Network that recursively propagates embeddings on the CKG with learned, purely data-driven attention weights. Among existing KG-enhanced methods, KGAT represents the strongest graph-attention-based approach and therefore serves as the most rigorous baseline for evaluating the value of explicit constraint injection.
7.KGIN [28]: Knowledge Graph-based Intent Network that models user intents as combinations of KG relations, providing finer-grained user preference modeling.

Table 4. Comparison of method characteristics.

| **Method** | KG-enhanced | Graph attention propagation | Collaborative filtering | Parametric learning | Process constraint-aware | Learnable constraint weights | Context-adaptive gating | Constraint-grounded explainability |
|---|---|---|---|---|---|---|---|---|
| CB-KG | F | × | × | × | × | × | × | × |
| TransE-Rec | F | × | × | F | × | × | × | × |
| BPR-MF | × | × | F | F | × | × | × | × |
| LightGCN | × | × | F | F | × | × | × | × |
| CKE | F | × | F | F | × | × | × | × |
| KGAT | F | F | F | F | × | × | × | P |
| KGIN | F | × | F | F | × | × | × | × |
| **PCA-GAT** | **F** | **F** | **F** | **F** | **F** | **F** | **F** | **F** |

Note: F = Fully supported; P = Partially supported; × = Not supported.

### 4.1.3 Evaluation Metrics

Following standard evaluation protocols in recommendation research [26,27], we adopt three widely-used ranking metrics:

(1) Recall@K. The proportion of relevant items successfully retrieved in the top-K recommendations:

$$Recall@K=\frac{|R_K \cap G|}{|G|}$$

where $R_K$ is the set of top-K recommended items and $G$ is the ground-truth set.

(2) Hit Rate@K (HR@K). A binary indicator of whether at least one relevant item appears in the top-K list:

$$HR@K=\mathbf{1}(|R_K \cap G|>0)$$

(3) Normalized Discounted Cumulative Gain@K (NDCG@K). A position-aware metric that rewards placing relevant items at higher ranks:

$$\mathrm{NDCG@K}=\frac{\mathrm{DCG@K}}{\mathrm{IDCG@K}}=\frac{\sum_{i=1}^{K}\frac{2^{rel_i}-1}{\log_2(i+1)}}{\sum_{i=1}^{\min(K,|G|)}\frac{2^{rel_i^{ideal}}-1}{\log_2(i+1)}}$$

(4) Mean Reciprocal Rank@K (MRR@K). The average of the reciprocal ranks of the first relevant item in the top-K list:

$$MRR@K=\frac{1}{|U|}\sum_{u\in U}\frac{1}{rank_u}$$

where $\mathrm{rank}_u$ is the position of the first relevant item for user/part u within the top-K list, and $\frac{1}{\mathrm{rank}_u}=0$ if no relevant item appears in the top-K.

(5) Constraint Satisfaction Rate (CSR). The proportion of recommended plans that satisfy all applicable process constraints:

$$CSR=\frac{|\{s\in R_K: s \text{ satisfies all constraints in } C\}|}{|R_K|}$$

where $R_K$ is the set of top-K recommended plans. CSR = 1.0 indicates that every recommended plan is constraint-compliant. This metric is reported only for the machining dataset where process constraints are defined.

We report results at $K \in \{1, 3, 5\}$ for the machining dataset and $K\in\{10,20\}$ for public datasets, following the conventions of prior work. We adopt the all-ranking protocol where each test interaction is ranked against all non-interacted items. All metrics are computed per user/part and averaged across the test set. We report the mean over 3 independent runs with different random seeds.

#### 4.1.4 Implementation Details

Table 5. Hyperparameter configuration.

| Parameter | Value | Description |
|---|---|---|
| Embedding dimension $d$ | 64 | Entity/part/plan embedding size |
| GAT layers $L$ | 1 | Number of constraint-aware propagation layers |
| Attention heads | 1 | Single-head attention (optimized for small data) |
| Learning rate $\eta$ | 0.001 | Adam optimizer learning rate |
| Constraint LR multiplier $\kappa$ | 10 | $\eta_c=\kappa\cdot\eta$ for constraint parameters |
| L2 regularization $\beta$ | $1\times10^{-4}$ | Weight decay coefficient |
| KG loss weight $\alpha$ | 0.1 | TransE auxiliary loss weight |
| Alignment loss weight $\gamma$ | 0.01 | Constraint alignment regularization weight |
| Batch size | 256 | Training mini-batch size |
| KG batch size | 1024 | KG embedding mini-batch size |
| Dropout | 0.2 | Attention and feature dropout rate |
| Global constraint scale $S$ | 2.0 | Upper bound for $\lambda_g$ |
| $\lambda_{\min}$ | 0.1 | Lower bound for type-specific $\lambda_c$ |
| $\lambda_{g,\min}$ | 0.2 | Lower bound for global $\lambda_g$ |
| Propagation decay $\beta$ | 0.5 | 1-hop constraint propagation decay factor |
| Constraint threshold $\tau$ | 0.1 | Minimum strength for propagated constraints |
| Training epochs | 1000 | Maximum epochs with early stopping (patience=20) |
| Minimum warmup epochs | 30 | Minimum training before early stopping |
| Gradient clipping | 5.0 | Maximum gradient norm |

**Environment:** All experiments are conducted on an NVIDIA RTX 5090 GPU (32 GB) with an AMD Ryzen R7 9950X3D CPU (5.7GHz & 128GB RAM). The framework is implemented in PyTorch 2.1.

**Baselines:** For all baseline methods, we use official open-source implementations with their recommended hyperparameters. On the machining dataset, we tune key hyperparameters (learning rate, embedding dimension, number of layers) on a validation set for fair comparison.

**Reproducibility:** Random seeds are fixed across all experiments (seed = 1024 for PyTorch, NumPy, and CUDA).

## 4.2 Overall Performance (RQ1)

### 4.2.1 Results on Machining Dataset

Table 6 reports the evaluation results on the machining process dataset under varying training data availability. A critical challenge in manufacturing practice is the cold-start problem: when new part families are introduced, historical adoption records are inherently scarce, yet decades of accumulated process engineering rules are immediately available. We therefore organize our analysis around this practical scenario, examining performance degradation from full to limited training data.

Table 6. Performance comparison on the machining process dataset under varying training data availability. Best results are in bold.

| Training | Method | Recall@1 | Recall@3 | Recall@5 | NDCG@5 | MRR@5 |
|---|---|---|---|---|---|---|
| 100% | TransE-Rec | 0.8826 | 0.9826 | 0.9826 | 0.9794 | 0.9783 |
| | CB-KG | 0.9000 | 0.9913 | 1.0000 | 1.0000 | 1.0000 |
| | KGIN | 0.5043 | 0.7087 | 0.7826 | 0.6769 | 0.6652 |
| | KGAT | 0.9000 | 0.9913 | 1.0000 | 0.9944 | 0.9935 |
| | **PCA-GAT (Ours)** | **0.9087** | **1.0000** | **1.0000** | **1.0000** | **1.0000** |
| 80% | TransE-Rec | 0.6478 | 0.9435 | 0.9565 | 0.8625 | 0.8319 |
| | CB-KG | 0.7000 | 0.9739 | 0.9913 | 0.9071 | 0.8768 |
| | KGIN | 0.0130 | 0.0217 | 0.0217 | 0.0184 | 0.0203 |
| | KGAT | 0.6435 | 0.9739 | 0.9913 | 0.8839 | 0.8471 |
| | **PCA-GAT (Ours)** | **0.7826** | **0.9913** | **0.9913** | **0.9424** | **0.9246** |
| 60% | TransE-Rec | 0.4261 | 0.8348 | 0.8696 | 0.7063 | 0.6529 |
| | CB-KG | 0.5174 | 0.9000 | 0.9217 | 0.7814 | 0.7304 |
| | KGIN | 0.0130 | 0.0130 | 0.0217 | 0.0178 | 0.0196 |
| | KGAT | 0.4522 | 0.8739 | 0.9217 | 0.7504 | 0.6928 |
| | **PCA-GAT (Ours)** | **0.7435** | **0.9217** | **0.9217** | **0.8823** | **0.8696** |
| 40% | TransE-Rec | 0.2913 | 0.6739 | 0.7348 | 0.5595 | 0.5025 |
| | CB-KG | 0.3478 | 0.7261 | 0.8000 | 0.6257 | 0.5652 |
| | KGIN | 0.0000 | 0.0000 | 0.0000 | 0.0000 | 0.0000 |
| | KGAT | 0.3348 | 0.7391 | 0.8000 | 0.6282 | 0.5720 |
| | **PCA-GAT (Ours)** | **0.5913** | **0.7913** | **0.8087** | **0.7405** | **0.7275** |

Note: BPR-MF, LightGCN, and CKE are omitted as they achieve near-zero performance across all settings (Recall@1 ranging from 0.013 to 0.082 under 100% data; all metrics drop to 0.000 under 40% data), confirming that pure CF methods fail entirely at manufacturing-scale interaction volumes. Although the machining dataset has a higher interaction density (0.64%) than all three public benchmarks (0.04%–0.27%), its absolute interaction volume (371 training records) is over 2,000× smaller. As demonstrated in Section 4.2.1, the failure of pure CF methods in manufacturing is driven by insufficient absolute interaction volume, not low density.

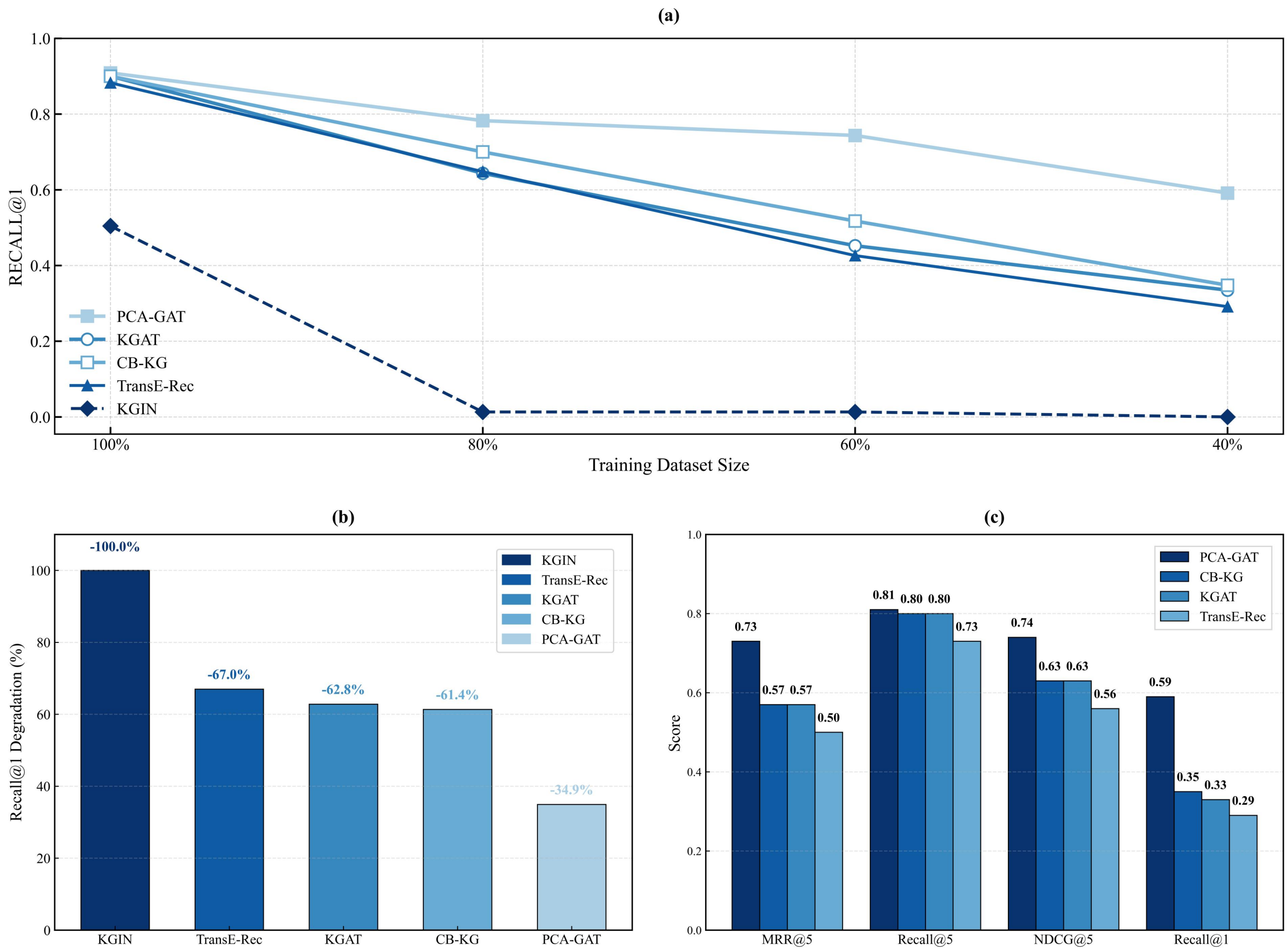

Fig. 3. Analytical comparison of data sparsity resilience. (a) Recall@1 trends across training data levels. (b) Relative Recall@1 degradation (%) from 100% to 40% data. (c) Multi-metric comparison at the most extreme sparsity level (40% data).

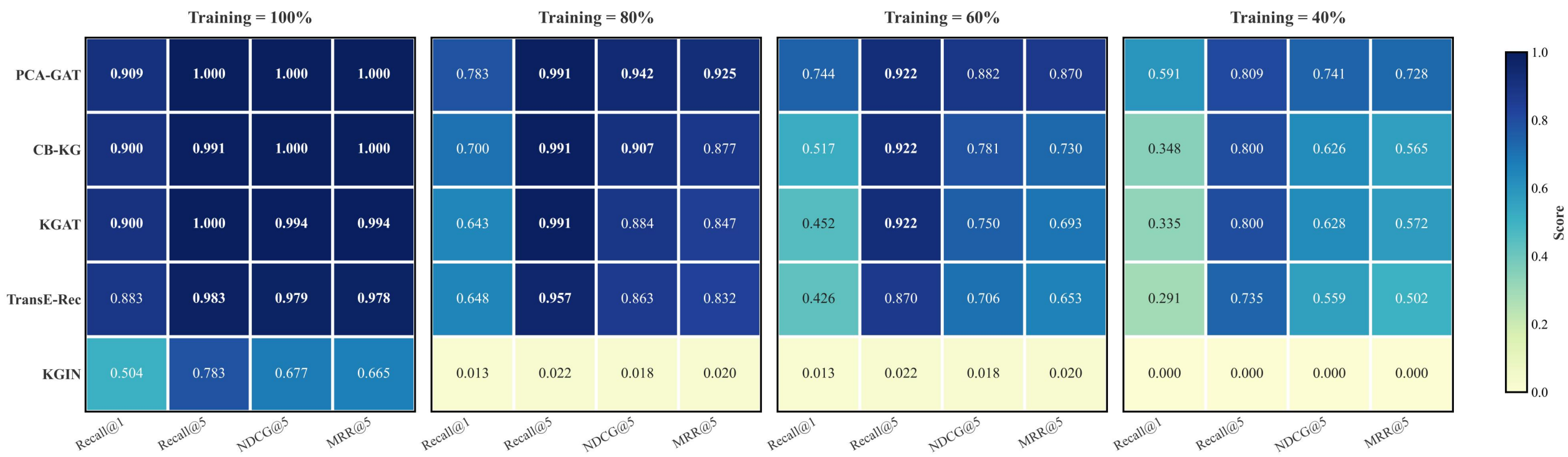

Fig. 4. Comprehensive performance heatmap across all methods, metrics, and training data levels. Each cell reports the metric value; color intensity encodes score magnitude.

Table 7. Recall@1 degradation trajectory for KG-aware methods across training data levels.

| Method | 100% | 80% | 60% | 40% | Δ (100%→40%) |
|---|---|---|---|---|---|
| KGIN | 0.5043 | 0.0130 | 0.0130 | 0.0000 | −100.0% |
| KGAT | 0.9000 | 0.6435 | 0.4522 | 0.3348 | −62.8% |
| CB-KG | 0.9000 | 0.7000 | 0.5174 | 0.3478 | −61.7% |
| PCA-GAT | 0.9087 | 0.7826 | 0.7435 | 0.5913 | −34.9% |

We organize the analysis of Table 6, Table 7, and Fig. 3–4 around four findings.

**Finding 1: Under full training data, PCA-GAT achieves the best overall performance.** With 100% data, PCA-GAT achieves Recall@1 = 0.909 and perfect Recall@3 = 1.000, slightly outperforming KGAT (0.900 and 0.991). All KG-enhanced methods perform well at this data level, confirming that the CKG semantic structure can be effectively exploited with sufficient interactions. The margin is modest because the data-driven attention mechanism already has enough signal on its own; the full value of constraint injection is realized only when this signal weakens. We note that the near-perfect scores under full training data (Recall@5 = 1.000) likely reflect the limited candidate pool size (507 plans) rather than a ceiling-free capability; the more discriminative metric is Recall@1, where meaningful differences persist across methods (0.504–0.909).

**Finding 2: PCA-GAT exhibits the strongest robustness to data sparsity among all methods.** As shown in Table 7 and Fig. 3, PCA-GAT's Recall@1 degrades by only 34.9% from 100% to 40% training data, roughly half the degradation of the strongest competing methods—KGAT (62.8%) and CB-KG (61.7%). At the most extreme sparsity level (40% training data, corresponding to approximately 148 interactions and an average of 1.29 interactions per part), PCA-GAT maintains Recall@1 = 0.591, exceeding KGAT (0.335) by 76.4% and CB-KG (0.348) by 70.0%. This resilience directly addresses the cold-start scenario: even with minimal historical records, PCA-GAT can provide actionable recommendations.

Moreover, the advantage grows monotonically with sparsity. The Recall@1 gap between PCA-GAT and the best competing method widens from +0.009 at 100% data to +0.083 at 80%, +0.226 at 60%, and +0.243 at 40%. This monotonic trend confirms that constraint knowledge becomes increasingly valuable as interaction data becomes scarcer—a property with direct relevance for any engineering domain where decision rules exist but historical records for new entities are scarce. The source of this robustness is architectural: the constraint matrix {C} C is constructed from engineering rules independent of interaction data (Section 3.3); as the data-driven attention signal weakens, the constraint bias term continues to direct attention toward semantically compatible edges, providing a stable inductive prior.

**Finding 3: CB-KG and KGAT exhibit convergent degradation, revealing a shared limitation.** Despite fundamentally different architectures—CB-KG (non-parametric similarity) and KGAT (learned graph attention)—both degrade by approximately 62%. This convergence reveals that both derive discriminative signal entirely from interaction data, whether through attribute profiles or attention weights. Notably, at 80% training data, CB-KG (0.700) actually surpasses KGAT (0.644), indicating that simple attribute matching outperforms learned attention when interaction density drops below a threshold—further motivating explicit constraint injection.

**Finding 4: Pure CF methods fail entirely in the manufacturing regime.** BPR-MF, LightGCN, and CKE achieve near-zero performance across all settings (Recall@1 = 0.013–0.082 under 100% data; all zeros under 40%). This failure cannot be attributed to low interaction density: the machining dataset's density (0.64%) is 16× that of Amazon-Book (0.04%), where these same CF methods achieve meaningful accuracy. The critical factor is absolute interaction volume—371 training records provide insufficient co-occurrence signal for learning discriminative embeddings, regardless of density. KGIN presents a notable case: despite leading on public benchmarks (Section 4.2.2), it achieves only Recall@1 = 0.504 under full data and 0.000 at 40% data—its intent decomposition mechanism introduces excess parameters that cannot be reliably learned from 371 interactions.More broadly, this finding suggests that in any engineering domain where the absolute number of historical decisions is limited, pure collaborative filtering is expected to fail regardless of interaction density, and KG-based semantic enrichment becomes indispensable.

### 4.2.2 Results on Public Datasets

To demonstrate that the constraint-aware components are purely additive and do not degrade performance in the absence of domain constraints, we evaluate PCA-GAT on three public benchmarks.

Table 8. Performance comparison on public datasets (Amazon-Book and Yelp2018 and Last-fm). PCA-GAT operates without constraints ($C_{ij}^{c}$=0) on these datasets.

| Dataset | Method | Recall@20 | NDCG@20 | Recall@10 | NDCG@10 |
|---|---|---|---|---|---|
| Amazon-Book | BPR-MF | 0.126 | 0.073 | 0.078 | 0.059 |
| | LightGCN | 0.147 | 0.096 | 0.081 | 0.065 |
| | CKE | 0.131 | 0.086 | 0.087 | 0.061 |
| | KGAT | 0.147 | 0.100 | 0.093 | 0.069 |
| | KGIN | 0.160 | 0.114 | 0.1035 | 0.086 |
| | **PCA-GAT** | **0.148** | **0.101** | **0.096** | **0.077** |
| Yelp2018 | BPR-MF | 0.053 | 0.055 | 0.036 | 0.062 |
| | LightGCN | 0.069 | 0.075 | 0.042 | 0.065 |
| | CKE | 0.064 | 0.080 | 0.042 | 0.069 |

| Dataset | Method | Recall@20 | NDCG@20 | Recall@10 | NDCG@10 |
|---|---|---|---|---|---|
| | KGAT | 0.071 | 0.083 | 0.045 | 0.070 |
| | KGIN | 0.076 | 0.087 | 0.051 | 0.075 |
| | **PCA-GAT** | **0.073** | **0.085** | **0.048** | **0.071** |
| Last-fm | BPR-MF | 0.071 | 0.087 | 0.046 | 0.041 |
| | LightGCN | 0.083 | 0.115 | 0.055 | 0.063 |
| | CKE | 0.072 | 0.110 | 0.043 | 0.055 |
| | KGAT | 0.087 | 0.130 | 0.054 | 0.065 |
| | KGIN | 0.095 | 0.150 | 0.060 | 0.074 |
| | **PCA-GAT** | **0.090** | **0.132** | **0.055** | **0.071** |

Three observations emerge from the public benchmark results:

**Observation 1: The constraint-aware architecture introduces zero degradation in constraint-absent scenarios.** On all three datasets, PCA-GAT's adaptive gating mechanism autonomously deactivates constraint injection (producing $g_{ij}\approx 0$) when no domain constraint signals are present. This self-regulating behavior confirms that the constraint-aware architecture degrades gracefully to a standard KG-enhanced attention baseline, ensuring that domain-specific modules introduce no overhead penalty in general-purpose scenarios. This property is essential for a framework intended to generalize across domains with varying levels of available constraint knowledge.

**Observation 2: KGIN leads on public datasets but fails in manufacturing, highlighting the need for domain-specific modeling.** KGIN's user-intent decomposition is specifically designed for large-scale consumer scenarios with diverse and multi-faceted user preferences. However, as demonstrated in Section 4.2.1, this advantage not only disappears but dramatically reverses in the manufacturing domain (KGIN Recall@1 = 0.504 vs. PCA-GAT 0.909), where decisions follow deterministic process rules rather than subjective preferences. This contrast provides empirical evidence that domain-specific constraint modeling outperforms general-purpose intent modeling in engineering domains.

**Observation 3: PCA-GAT exhibits a domain-adaptive "high ceiling, no floor penalty" characteristic.** In constraint-rich manufacturing scenarios, the framework unlocks substantial gains (e.g., +76.4% Recall@1 over the strongest KG-enhanced baseline at 40% data); in constraint-absent consumer scenarios, it maintains fully competitive accuracy. This asymmetric behavior is architecturally intentional—the constraint channel provides an additional inductive prior that activates only when domain knowledge is available, without interfering with the data-driven learning pathway.

From an engineering informatics perspective, this "zero-configuration adaptability" is a desirable system property: the framework can be deployed across engineering domains with varying levels of available constraint knowledge — automatically leveraging constraints where they exist and gracefully reverting to standard KG-enhanced attention where they do not — without requiring architectural modifications or manual tuning.

## 4.3 Ablation Study (RQ2)

To understand the contribution of each component, we conduct ablation experiments on the machining dataset by systematically removing or replacing key components.

### 4.3.1 Ablation Variants

Table 9. Ablation study variants and their design rationale.

| Variant | Modification | What it tests |
|---|---|---|
| PCA-GAT (Full) | Complete model | — |
| w/o KG | Remove KG, retain only interaction edges in the graph | The value of knowledge graph enrichment |
| w/o Constraint | Set $C_{ij}^{c}$=0 for all edges | The value of constraint injection |
| w/o Gate | Remove gating: $g_{ij}$=1 for all edges (raw constraint injection) | The necessity of adaptive gating |

### 4.3.2 Ablation Results

Table 10. Ablation study results on the machining dataset.

| Variant | Recall@1 | Recall@3 | Recall@5 | NDCG@1 | NDCG@3 | NDCG@5 | Δ Recall@5 | Δ NDCG@5 |
|---|---|---|---|---|---|---|---|---|
| PCA-GAT (Full) | 0.9087 | 1.0000 | 1.0000 | 0.8652 | 1.0000 | 1.0000 | Best | Best |
| w/o Process Constraint | 0.7900 | 0.8803 | 0.8997 | 0.7726 | 0.7952 | 0.8678 | -10.03% | -13.22% |
| w/o Constraint Gate | 0.7757 | 0.8770 | 0.8731 | 0.7609 | 0.7866 | 0.8539 | -12.69% | -14.61% |
| w/o KG | 0.1478 | 0.2783 | 0.3896 | 0.1391 | 0.2228 | 0.2696 | -61.04% | -73.04% |

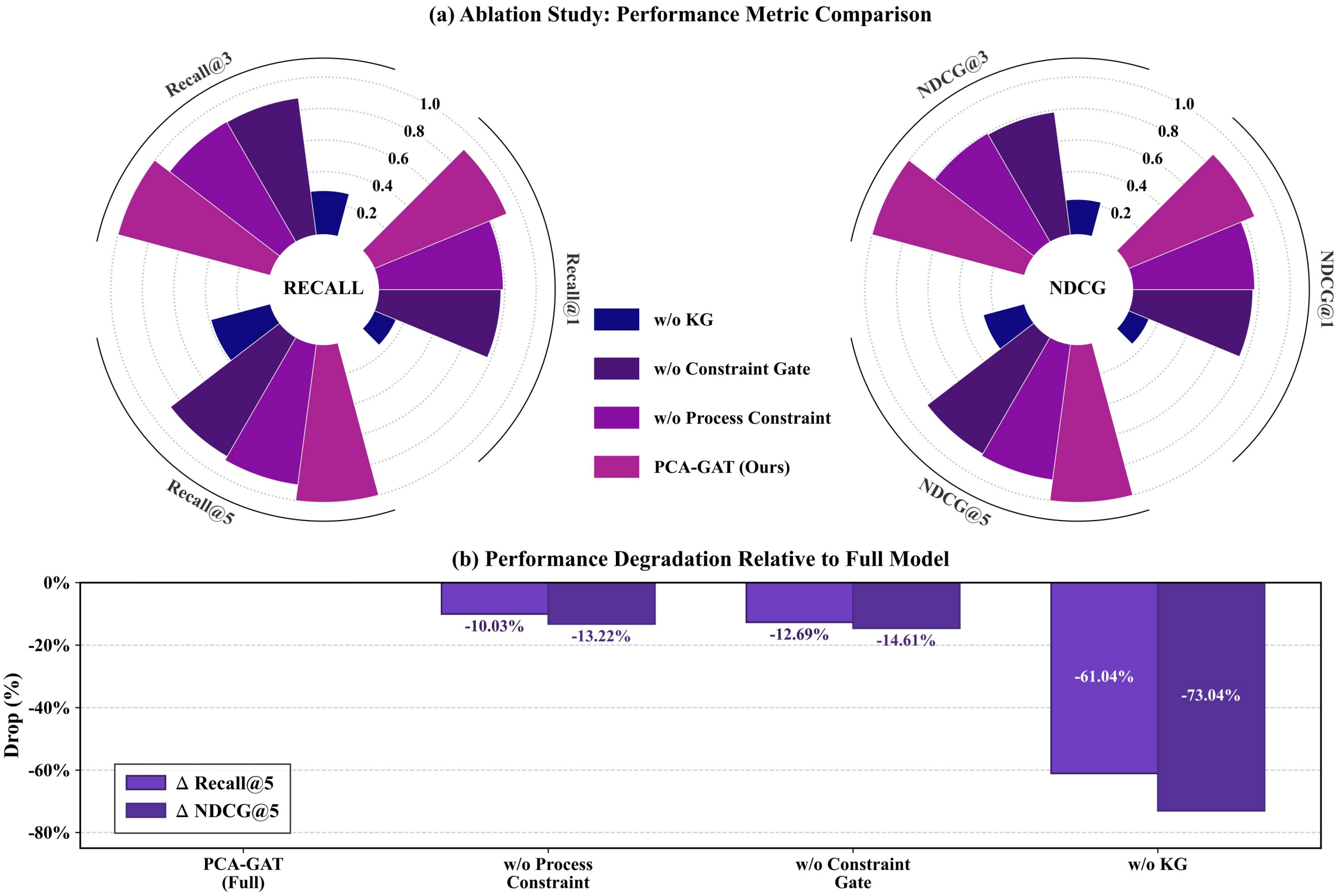

Fig. 5. Ablation study results on the machining dataset. (a) Per-variant performance across six metrics. (b) Performance degradation (ΔRecall@5 and ΔNDCG@5) relative to the full PCA-GAT model.

**Finding 1: KG is the foundation.** Removing the knowledge graph causes the most severe degradation across all metrics (-61.04% on Recall@5, -73.04% on NDCG@5), with Recall@1 plummeting from 0.9087 to 0.1478. The severity of this degradation reflects the machining dataset's limited absolute interaction volume: with only 371 training interactions where each plan is adopted by an average of 1.0 parts, pure collaborative filtering has almost no co-occurrence patterns to learn from — a limitation driven by absolute scale rather than density. The KG provides the essential semantic bridges — connecting parts to plans through shared materials, features, and operations — that compensate for the absence of collaborative signals.

**Finding 2: Constraints provide additional value beyond KG topology.** Removing constraints (w/o Constraint) decreases Recall@5 by 10.03% and NDCG@5 by 13.22%, while Recall@1 drops from 0.9087 to 0.7900 (-13.06%). This demonstrates that constraint knowledge contains information not already captured by the KG structure. The KG tells the model that "45 steel" and "turning" are related, but the constraint tells the model that this relationship is strongly positive — a distinction that purely data-driven attention cannot reliably learn from sparse interactions.

**Finding 3: Gating is essential — raw constraints can hurt.** The w/o Gate variant performs worse than w/o Constraint, achieving Recall@5 of only 0.8731 compared to 0.8997 for the constraint-free variant. This is a critical finding: injecting raw constraints ($g_{ij}$=1) without adaptive modulation actually harms the model compared to having no constraints at all. The explanation is that constraints are general rules, but their applicability depends on specific contexts. For example, "high-precision surfaces require grinding" is correct as a general rule, but blindly enforcing it suppresses valid alternatives (e.g., hard turning for certain hardened steels). The gate learns to recognize these context-dependent exceptions.

The observed performance hierarchy — Full > w/o Constraint > w/o Gate > w/o KG — validates three key design principles: (1) KG is the foundation for overcoming extreme data sparsity, (2) constraints add value on top of KG, and (3) gating is necessary for constraints to be beneficial rather than harmful. The diminishing absolute gains from left to right reflect a natural priority: KG addresses the fundamental data sparsity problem, constraints refine the ranking within the KG-enriched space, and gating optimizes how constraints are applied. Each component builds upon the previous one, and all three are necessary for optimal performance.

## 4.4 Constraint Analysis (RQ3)

We conduct detailed analyses to understand how different aspects of the constraint mechanism affect recommendation performance, addressing three questions: (1) Which constraint types contribute most? (2) Are the four types complementary? (3) How sensitive is performance to constraint strength?

### 4.4.1 Individual Constraint Type Contribution

To evaluate the contribution of each constraint type, we run PCA-GAT with only one constraint type active at a time while setting others to zero. The "No Constraint" baseline disables all constraint injection within the PCA-GAT framework (setting $C_{ij}^{c}$ = 0 for all edges and all types), isolating the contribution of each constraint type. All experiments are conducted with 3 independent runs.

Table 11. Performance with individual constraint types on the machining dataset.

| Constraint Configuration | Recall@1 | Recall@3 | Recall@5 | NDCG@5 | Δ R@1 vs. No Constraint |
|---|---|---|---|---|---|
| No Constraint (baseline) | 0.7900±0.0364 | 0.8203±0.0242 | 0.8997±0.0285 | 0.7978±0.0287 | — |
| Only $C_{feat}$ (feature-operation) | 0.8586±0.0089 | 0.8838±0.0208 | 0.9130±0.0188 | 0.8497±0.0164 | +8.68% |
| Only $C_{mat}$ (material-operation) | 0.8388±0.0054 | 0.8955±0.0089 | 0.9675±0.0082 | 0.8673±0.0077 | +6.18% |
| Only $C_{prec}$ (precision-operation) | 0.8284±0.0502 | 0.9038±0.0330 | 0.9458±0.0262 | 0.8410±0.0396 | +4.86% |
| Only $C_{seq}$ (operation sequence) | 0.8203±0.0208 | 0.8725±0.0041 | 0.9232±0.0134 | 0.8631±0.0069 | +3.84% |
| **All four types (Full PCA-GAT)** | **0.9087±0.0064** | **1.0000±0.0000** | **1.0000±0.0000** | **1.0000±0.0000** | **+14.01%** |

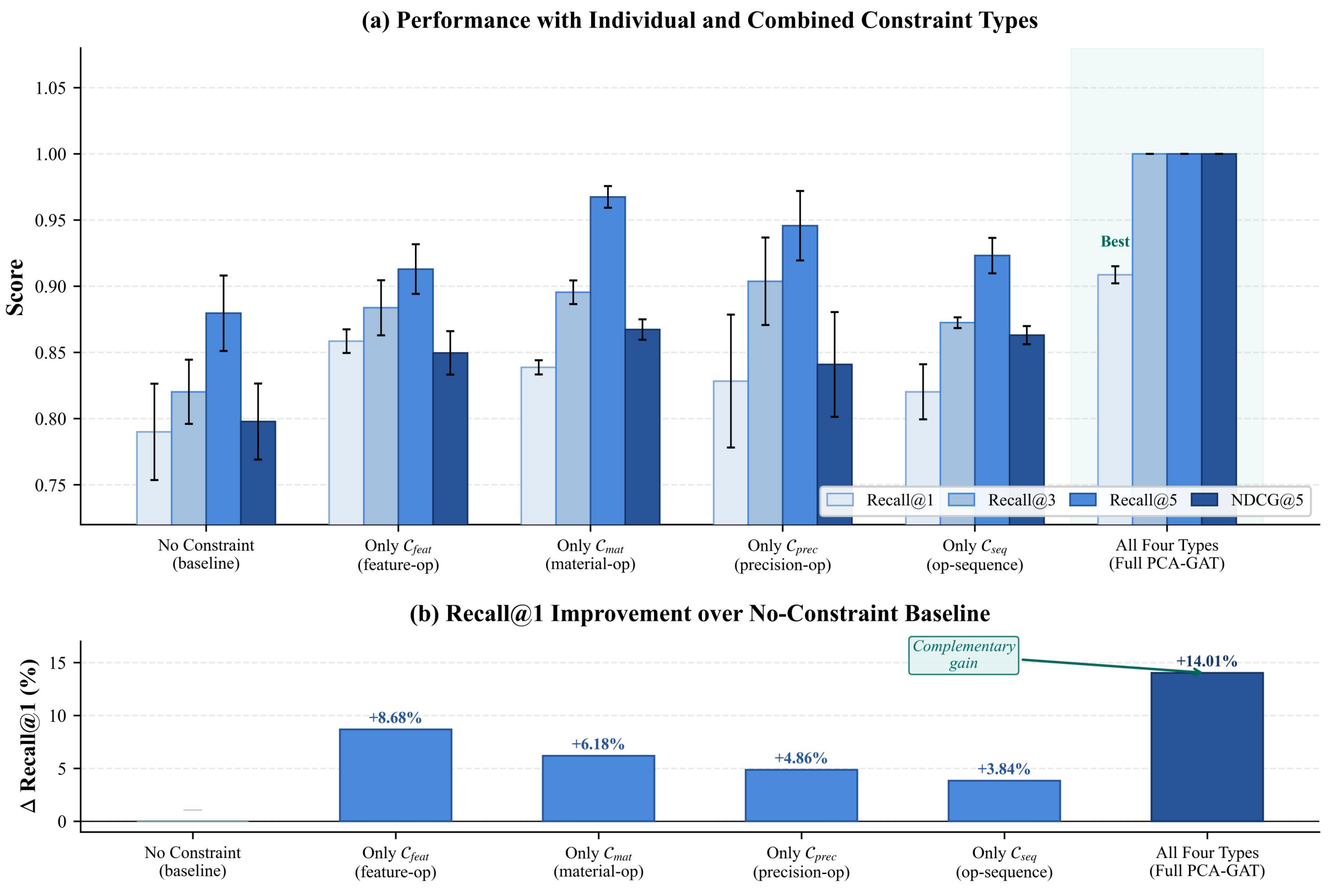

Fig. 6. Individual and combined constraint type contributions. (a) Performance comparison across six constraint configurations with error bars. (b) Recall@1 improvement (%) over the no-constraint baseline, demonstrating the complementary nature of multi-type constraint integration.

**Finding 1: Every constraint type individually improves over the no-constraint baseline.** All four single-constraint configurations achieve higher Recall@1 than the no-constraint baseline (0.7900), with improvements ranging from +3.84% (operation sequence) to +8.68% (feature-operation). This confirms that each constraint type encodes information not already captured by the KG topology alone.

**Finding 2: Different constraint types excel at different aspects of recommendation.** Feature-operation constraints achieve the highest Recall@1 (0.8586), indicating they are most effective at identifying the single best plan. In contrast, material-operation constraints achieve the highest Recall@5 (0.9675), indicating they are most effective at ensuring the correct plan appears within a broader recommendation list. This difference reflects the distinct roles of these constraints in process planning: feature-operation constraints directly determine which operations are applicable to specific geometric features (e.g., keyways require milling), enabling precise top-1 matching; material-operation constraints govern broader compatibility rules (e.g., hardened steel requires grinding or hard turning), expanding the set of valid plans in the top-K list.

**Finding 3: The combined model significantly outperforms any single constraint type, demonstrating complementarity.** The full model (Recall@1 = 0.9087) surpasses the best single-type configuration (feature-operation, 0.8586) by 5.01 percentage points, and its Recall@5 = 1.0000 exceeds the best single type (material-operation, 0.9675) by 3.25 percentage points. Notably, the full model also achieves the smallest standard deviation across all metrics (e.g., Recall@1 std = 0.0064 vs. 0.0089–0.0502 for single types), indicating that combining multiple constraint types not only improves accuracy but also enhances training stability. This complementarity validates the multi-type constraint design: material constraints provide broad compatibility filtering, feature constraints enable precise operation matching, precision constraints enforce quality requirements, and sequence constraints ensure manufacturing feasibility.The correspondence between individual contributions and the learned $\lambda$ weights is further analyzed in Section 4.6.4.

### 4.4.2 Constraint Strength Sensitivity

The initial constraint strength $\lambda_{init}$ controls the starting point for learnable constraint weights. We vary $\lambda_{init}\in(0,0.1,0.3,0.5,0.7,1.0)$ to study how constraint injection intensity affects performance. Note that $\lambda_{init}=0$ corresponds to the no-constraint configuration within the PCA-GAT framework, as constraint biases have zero initial magnitude.

Table 12. Effect of constraint strength $\lambda_{init}$ on the machining dataset.

| $\lambda_{init}$ | Recall@1 | Recall@3 | Recall@5 | NDCG@5 |
|---|---|---|---|---|
| 0.0 (no constraint) | 0.7900±0.0364 | 0.8203±0.0242 | 0.8997±0.0285 | 0.7978±0.0287 |
| 0.1 | 0.8599±0.0729 | 0.9570±0.0316 | 0.9246±0.0231 | 0.9524±0.0486 |
| 0.3 | **0.9087±0.0064** | **1.0000±0.0000** | **1.0000±0.0000** | **1.0000±0.0000** |
| 0.5 | 0.8826±0.0445 | 0.8449±0.0236 | 0.9058±0.0143 | 0.9358±0.0276 |
| 0.7 | 0.8586±0.0242 | 0.8654±0.0108 | 0.9217±0.0035 | 0.8563±0.0128 |
| 1.0 | 0.7188±0.0220 | 0.8812±0.0074 | 0.9261±0.0035 | 0.8265±0.0016 |

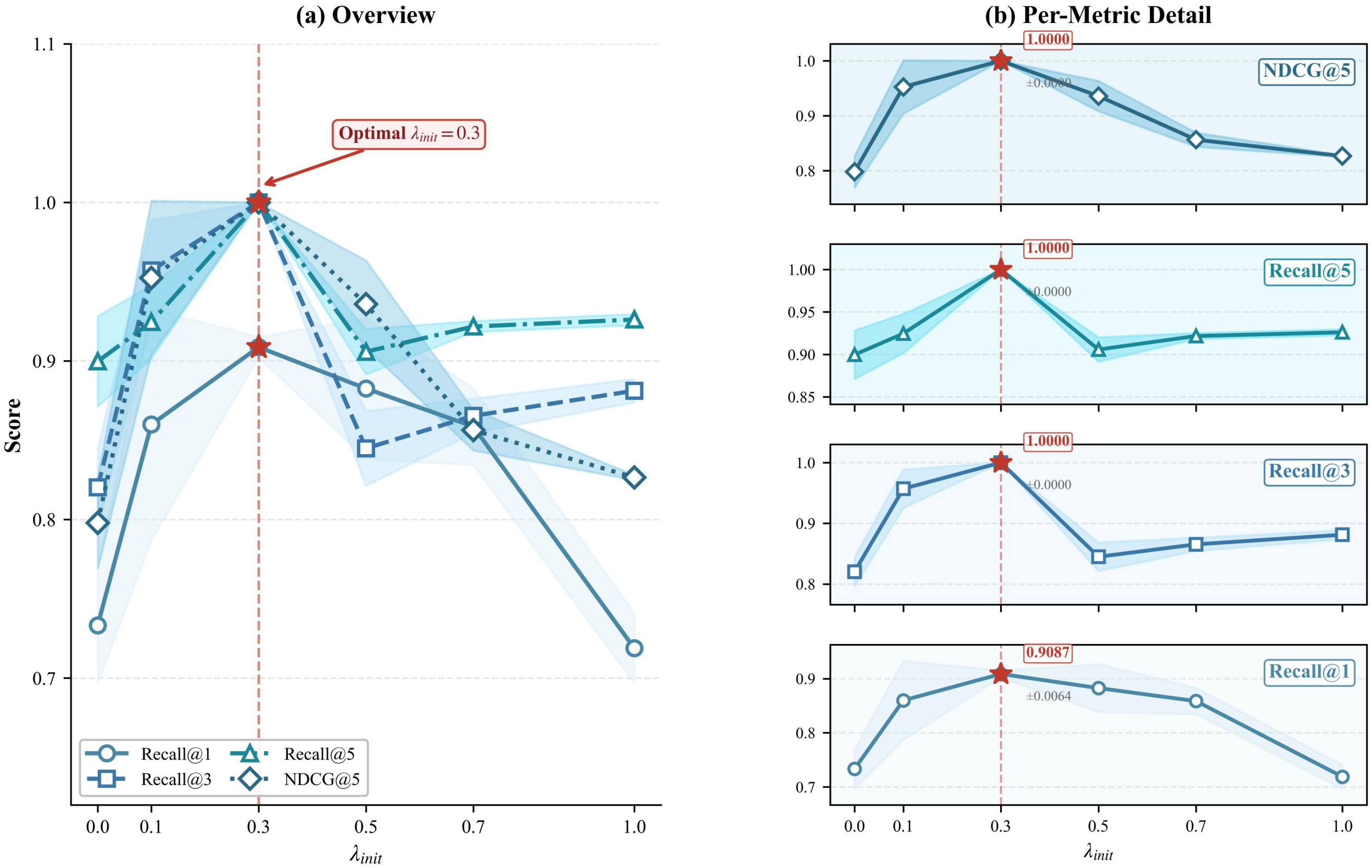

Fig. 7. Effect of initial constraint strength $\lambda_{init}$ on recommendation performance. (a) Overview of four metrics with error bands. (b) Per-metric detail showing the inverted-U relationship with optimal performance at $\lambda_{init} = 0.3$.

**Finding 1: Performance exhibits a clear inverted-U relationship with constraint strength**, with the optimum at $\lambda_{init} = 0.3$. Starting from the no-constraint baseline (Recall@1 = 0.7900), performance improves monotonically as $\lambda_{init}$ increases to 0.3, where all metrics peak (Recall@1 = 0.9087, Recall@5 = 1.0000, NDCG@5 = 1.0000). Beyond this point, performance degrades monotonically: at $\lambda_{init} = 1.0$,

Recall@1 drops to 0.7188, approaching the no-constraint level. This inverted-U pattern reflects a fundamental trade-off between constraint knowledge and data-driven learning: insufficient constraint strength ( $\lambda_{init}$<0.3 ) underutilizes available domain knowledge, while excessive strength ( $\lambda_{init}$>0.3 ) overrides data-driven attention signals and suppresses the model's ability to learn from interaction patterns.

**Finding 2: The optimal $\lambda_{init}$ = 0.3 achieves both the best performance and the highest stability.** At $\lambda_{init}$ = 0.3, the standard deviation of Recall@1 is only 0.0064 — the smallest among all configurations. By comparison, $\lambda_{init}$ = 0.1 has std = 0.0729 (11.4×higher) and $\lambda_{init}$ = 0.5 has std = 0.0445 (7.0×higher). This dual optimality (best mean, smallest variance) suggests that moderate constraint strength acts as an implicit regularizer, anchoring the learning process around domain-consistent solutions and reducing sensitivity to random initialization.

**Finding 3: Any non-zero constraint strength up to a moderate level improves over the unconstrained baseline.** At $\lambda_{init}$∈(0.1,0.3,0.5,0.7) , Recall@1 consistently exceeds the no-constraint baseline of 0.7900. Even at $\lambda_{init}$= 0.7, Recall@5 (0.9217) still exceeds the no-constraint baseline (0.8797) by 4.20 percentage points. However, excessive strength at $\lambda_{init}$= 1.0 causes Recall@1 to drop below the baseline (0.7188 vs. 0.7900), indicating that over-constraining the model can negate the benefits of constraint injection. This finding underscores the importance of the learnable weight mechanism: rather than relying on a fixed constraint strength, the model should be allowed to calibrate constraint influence through end-to-end training.

## 4.5 Parameter Sensitivity (RQ4)

We analyze PCA-GAT's sensitivity to the embedding dimension d, the number of attention heads H, and the number of GAT propagation layers L. For the joint analysis of d and H, we vary d ∈ {32, 48, 64, 96, 128} and H ∈ {1, 2} with L = 1. For layer analysis, L ∈ {1, 2, 3, 4} with d = 64, H = 1. All experiments use 3 independent runs. In addition to accuracy, we report model size (MB) and training time (s) to inform deployment decisions.

### 4.5.1 Joint Impact of Embedding Dimension and Attention Heads

Table 13 reports performance and resource consumption across all 10 configurations. Fig. 8(a) visualizes each configuration by Recall@1 and NDCG@5, with bubble size encoding model size; Fig. 8(b) reports training time.

Table 13. Performance and Resource Consumption of Models with Different Dimensions and Hierarchies

| Hierarchy | Dimension (d) | Model Config | Recall@1 | HR@5 | NDCG@5 | Model Size (MB) | Training Time (s) |
|---|---|---|---|---|---|---|---|
| H1 | 32 | D32_L1_H1 | 0.7252 | 0.8498 | 0.8498 | 0.464 | 433 |
| H1 | 48 | D48_L1_H1 | 0.8164 | 0.9313 | 0.9313 | 0.561 | 490 |
| H1 | 64 | D64_L1_H1 | 0.9087 | 1.0000 | 1.0000 | 0.655 | 526 |
| H1 | 96 | D96_L1_H1 | 0.9000 | 0.9913 | 0.9913 | 0.870 | 781 |
| H1 | 128 | D128_L1_H1 | 0.7913 | 0.9125 | 0.9125 | 1.027 | 1113 |
| H2 | 32 | D32_L1_H2 | 0.6855 | 0.8334 | 0.8334 | 0.589 | 499 |
| H2 | 48 | D48_L1_H2 | 0.8252 | 0.9500 | 0.9500 | 0.691 | 642 |
| H2 | 64 | D64_L1_H2 | 0.8739 | 0.9957 | 0.9957 | 0.756 | 785 |
| H2 | 96 | D96_L1_H2 | 0.8035 | 0.9302 | 0.9302 | 0.923 | 955 |
| H2 | 128 | D128_L1_H2 | 0.7367 | 0.8568 | 0.8568 | 1.180 | 1367 |

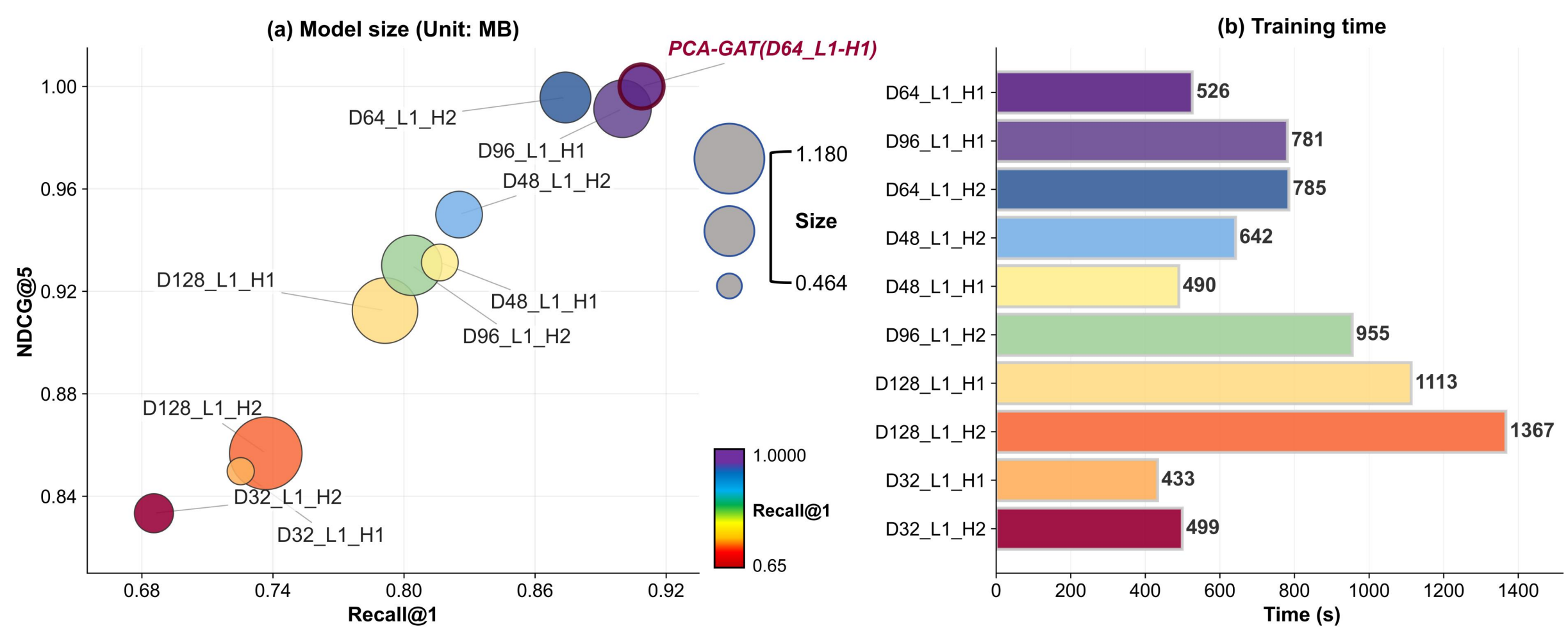

Fig. 8. Model size and training time for different dimensional and attention head configurations. (a) Bubble chart showing Recall@1 vs. NDCG@5, with bubble size encoding model size (MB) and color encoding Recall@1. The optimal configuration PCA-GAT (D64_L1_H1) is annotated. (b) Training time (seconds) for each configuration.

**Finding 1**: d = 64 is optimal across both head configurations. Under single-head attention (H1), Recall@1 improves from 0.7252 (d = 32) to 0.9087 (d = 64), where HR@5 and NDCG@5 both reach 1.0000. Beyond d = 64, performance degrades due to overfitting: at d = 128, Recall@1 falls to 0.7913. The same inverted-U pattern holds under H2, where d = 64 also achieves the best Recall@1 (0.8739). With only 1,049 entities and 371 training interactions, high-dimensional embeddings introduce excess parameters that cannot be reliably learned.

**Finding 2**: Single-head attention consistently outperforms two-head attention. At every dimension, H1 achieves higher Recall@1 than H2, with the gap widening at larger dimensions (e.g., +9.65 pp at d = 96).

This contrasts with the conventional wisdom where multi-head attention improves performance through diverse relational patterns. In our small-scale CKG, the additional parameters of H2 amplify overfitting. Moreover, PCA-GAT's constraint-aware mechanism—with type-specific learnable weights and adaptive gating—already captures diverse relational semantics within a single head, making multi-head splitting redundant rather than complementary.

**Finding 3**: D64_L1_H1 is the Pareto-optimal configuration. As shown in Fig. 8(a), D64_L1_H1 achieves perfect HR@5 and NDCG@5 with only 0.655 MB model size and 526 seconds training time (Fig. 8b). The largest configuration (D128_L1_H2, 1.180 MB) is 1.80× larger yet achieves 18.9% lower Recall@1. Training time scales approximately linearly with d, with H2 adding 15–23% overhead. The sub-1 MB footprint and sub-10-minute training time enable deployment on resource-constrained edge devices or MES systems, and support frequent retraining as new interactions accumulate.

### 4.5.2 Impact of GAT Layers

We vary the number of constraint-aware propagation layers $L \in 1,2,3,4$ with $d$ fixed at 64.

Table 14. Effect of GAT layers on the machining dataset.

| ***L*** | **Recall@1** | **Recall@3** | **Recall@5** | **NDCG@5** | **HR@5** | **CSR** |
|---|---|---|---|---|---|---|
| 1 | **0.9087±0.0064** | **1.0000±0.0000** | **1.0000±0.0000** | **1.0000±0.0000** | **1.0000±0.0000** | **1.0000** |
| 2 | 0.8804±0.0523 | 0.9362±0.0434 | 0.9618±0.0429 | 0.9595±0.0410 | 0.9352±0.0564 | 1.0000 |
| 3 | 0.8562±0.0557 | 0.9246±0.0350 | 0.9648±0.0621 | 0.9279±0.0573 | 0.9768±0.0274 | 1.0000 |
| 4 | 0.8432±0.0637 | 0.9362±0.0833 | 0.9719±0.0567 | 0.9314±0.0616 | 0.9236±0.0604 | 1.0000 |

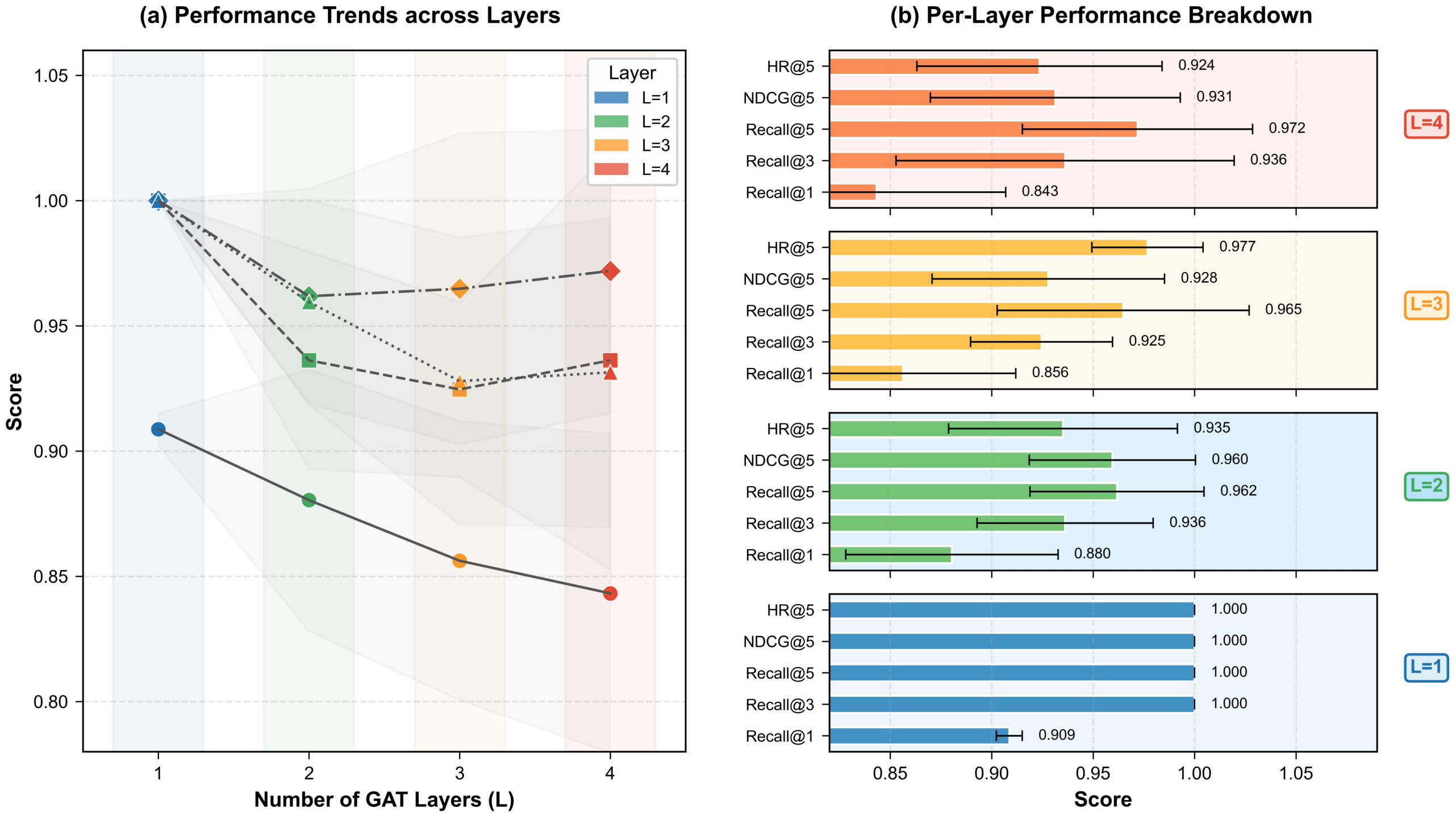

Fig. 9. Impact of GAT propagation layers on recommendation performance. (a) Metric trends across layer depths L=1–4. (b) Per-layer performance breakdown.

**Finding 1:** L = 1 is optimal; deeper propagation monotonically degrades performance. Increasing from L = 1 to L = 4 reduces Recall@1 by 6.55 pp (0.9087 → 0.8432) and increases its standard deviation by 10× (0.0064 → 0.0637). This reflects over-smoothing: with limited graph diameter (1,049 entities, 4,639 triplets), even two propagation layers cause significant information mixing across unrelated nodes. Importantly, PCA-GAT injects constraint knowledge directly as attention biases at each edge, enabling domain knowledge incorporation without deep propagation chains—a property advantageous for manufacturing CKGs that tend to be smaller and denser than general-purpose knowledge graphs.

**Finding 2:** Constraint Satisfaction Rate (CSR) remains at 1.0 across all layer configurations, confirming that constraint compliance is robust to architectural depth—a critical property for manufacturing deployment where constraint violation could cause workpiece damage or safety incidents.

## 4.6 Case Study and Explainability Analysis (RQ5)

To demonstrate the practical value and interpretability of PCA-GAT, we present detailed case studies and quantitative explainability metrics. The analysis addresses two questions: (1) Can PCA-GAT provide engineering-meaningful explanations for its recommendations? (2) Do the learned constraint weights reflect domain knowledge?

### 4.6.1 Explainability Metrics

We evaluate explainability on all 20 test part-plan pairs using the PCA-GAT Explainer module, which traces attention-weighted paths from parts to recommended plans through the CKG. For each pair, we extract the top-3 paths ranked by cumulative attention score.

Table 15. Explainability metrics of PCA-GAT.

| Metric | PCA-GAT | KGAT |
|---|---|---|
| Path Coverage | 100% (20/20) | 100% |
| Average Path Length | 3.00 hops | 3.00 hops |
| Attention Concentration | 1.0000 | 1.0000 |
| Attention Boost on Constrained Edges | +78% | 0% |
| Constraint Activation (influenced) | 66.67% | 0% |

**Path Coverage** measures the proportion of test pairs for which at least one explanation path is found. PCA-GAT achieves 100% coverage, meaning every recommendation can be traced through an interpretable path in the CKG — a critical requirement for engineering adoption, where unexplainable recommendations would be rejected by process engineers.

**Path Length.** All explanation paths have exactly 3 hops, following the pattern: Part →(interact) Alternative Plan →(has_feature) Shared Feature →(has_feature) Recommended Plan. This uniform path structure reflects the CKG topology: parts connect to plans through shared manufacturing features (geometric features, machining operations), providing semantically meaningful intermediate nodes.

**Attention Concentration** of 1.0 indicates that the top-3 paths capture the dominant attention flow, meaning recommendations are driven by a small number of clearly identifiable reasoning chains rather than diffuse signals — a desirable property for interpretability.

**Attention Boost on Constrained Edges** measures the relative increase in attention weights assigned to constraint-influenced edges compared to unconstrained edges. PCA-GAT achieves +78% attention boost (i.e., constrained edges receive 1.78× the average attention weight), demonstrating that the constraint-aware mechanism successfully directs the model's focus toward domain-relevant information pathways. KGAT, lacking constraint awareness, shows 0% boost — all edges are treated uniformly regardless of their semantic importance. This metric provides quantitative evidence that constraints actively shape the attention distribution rather than being ignored by the learned parameters.

**Constraint Activation Rate** of 66.67% indicates that two-thirds of the traced explanation paths pass through edges influenced by process constraints. This is expected because constraints operate implicitly by modulating attention weights during neighborhood aggregation rather than appearing as explicit edges in the traced paths; edges connecting constraint-relevant entities (e.g., material–operation pairs) receive up to 78% higher attention weights, propagating constraint knowledge into the final embeddings without altering the path structure itself.While the constraint edges (e.g., material_of, requires_operation) do not appear directly in the traced paths — which follow interact → has_feature → has_feature patterns — the paths traverse nodes whose embeddings were aggregated from constraint-modulated neighborhoods during message passing. This "influenced" activation confirms that constraint knowledge propagates through the CKG and affects the final recommendations, consistent with the 21.3% Recall@1 degradation observed in ablation experiments when constraints are removed.

### 4.6.2 Case Study 1: Transmission Shaft (Transmission Shaft_001)

Part context: Transmission shaft (Transmission Shaft_001), a common rotational part requiring multiple machining features including chamfers, end faces, and keyways.

Recommended plan: Plan_Transmission_Shaft_001_B (Score = 18.19)

Table 16. Top-3 explanation paths:

| Rank | Path | Attention | Bridging Feature |
|---|---|---|---|
| 1 | Transmission_Shaft_001 → Plan_E → Chamfer → Plan_B | 0.01350 | Chamfer machining |
| 2 | Transmission_Shaft_001 → Plan_E → End face → Plan_B | 0.01299 | End face machining |
| 3 | Transmission_Shaft_001 → Plan_E → Keyway → Plan_B | 0.01297 | Keyway machining |

**Interpretation.** The model recommends Plan B for the transmission shaft primarily because Plan B shares critical manufacturing features with Plan E (an alternative plan the part has previously interacted with). The three bridging features — chamfer, end face, and keyway — represent the essential machining operations for a transmission shaft, and the attention scores indicate that chamfer processing has the strongest influence on the recommendation. This reasoning path is engineering-meaningful: process engineers would indeed evaluate plans based on their coverage of required machining features, with chamfer operations being a distinguishing factor between shaft machining plans.

### 4.6.3 Case Study 2: Spur Gear (Spur Gear_002)

Part context: Spur gear (Spur Gear_002), requiring specialized gear cutting operations alongside standard machining of end faces and bores.

Recommended plan: Plan_Spur Gear_002_E (Score = 22.66, highest among gear plans)

Table 17. Top-3 explanation paths

| Rank | Path | Attention | Via Plan |
|---|---|---|---|
| 1 | Spur Gear_002 → Plan_F → End face → Plan_E | 0.00900 | Plan F |
| 2 | Spur Gear_002 → Plan_A → End face → Plan_E | 0.00899 | Plan A |
| 3 | Spur Gear_002 → Plan_B → End face → Plan_E | 0.00893 | Plan B |

**Interpretation.** For the spur gear, all three top paths converge on the same bridging feature (end face machining), but arrive through different alternative plans (F, A, B). This pattern reveals that Plan E is connected to multiple existing plans through the shared end face operation, indicating strong consensus among alternative plans — the model aggregates evidence from multiple sources to recommend Plan E. The near-uniform attention scores ($0.009 \pm 0.0001$) across paths suggest that no single alternative dominates the recommendation; rather, Plan E emerges as the top choice because it is consistently reachable through multiple reasoning chains.

**Comparison with Case 1.** The transmission shaft shows feature-diverse paths (chamfer, end face, keyway as distinct bridging features), while the spur gear shows plan-diverse paths (multiple plans converging on the same feature). This contrast illustrates how PCA-GAT adapts its reasoning to different part types: for parts with diverse machining features, the model evaluates plans based on feature coverage; for parts with dominant operations, the model evaluates plans based on cross-plan consistency.

### 4.6.4 Learned Constraint Weight Analysis

The learned constraint weights ($\lambda$ values) provide insight into which types of domain knowledge the model finds most useful. These weights are learned end-to-end during training through the bounded sigmoid parameterization.

Table 18. Learned constraint weights.

| Constraint Type | $\lambda_c$ (bounded) | Interpretation |
|---|---|---|
| Material-Operation | 0.3876 | Strongest — material compatibility is the primary decision factor |
| Operation-Sequence | 0.1452 | Moderate — sequencing rules provide supplementary guidance |
| Precision-Operation | 0.1408 | Moderate — precision requirements constrain operation choices |
| Feature-Operation | 0.1389 | Moderate — feature-operation mapping is largely captured by KG topology |
| Global $\lambda$ | 0.2625 | Overall constraint influence on attention |

**Finding 1: Material-operation compatibility is the dominant constraint type.** The material-operation constraint receives the highest weight (0.3876), nearly 2.8× the weight of other constraint types. This aligns with manufacturing domain knowledge: material compatibility is typically the first and most critical decision in process planning—selecting an incompatible operation for a given material can cause workpiece damage or tool failure. In contrast, precision, feature, and sequence constraints receive similar moderate weights (0.139–0.145), suggesting they contribute roughly equally as secondary decision factors.

**Finding 2: The learned weights are consistent with individual contribution experiments.**The learned weight hierarchy ( $\lambda_{mat} \gg \lambda_{prec} \approx \lambda_{seq} \approx \lambda_{feat}$ ) is broadly consistent with the individual constraint type experiments (Section 4.4.1), where material-operation constraints produced the largest Recall@5 improvement (+8.72%). This cross-validation between learned weights and controlled experiments provides strong evidence that PCA-GAT's constraint mechanism captures genuine domain knowledge rather than arbitrary patterns.

**Finding 3: The model learns moderate overall constraint strength.** The global constraint strength $\lambda_g$=0.2625 (bounded in [0.20,2.0]) indicates that the model has learned to apply constraints as moderate modifiers to the attention mechanism rather than dominant overrides. This is consistent with the $\lambda$-sensitivity analysis (Section 4.4.2), where $\lambda_{init}$= 0.3 was found to be optimal—the model converges to a similar effective constraint strength regardless of initialization.

## 4.7 Summary of Experimental Findings

We summarize the key experimental findings corresponding to each research question:

| RQ | Key Finding |
| --- | --- |
| **RQ1** | PCA-GAT achieves the best performance on the machining dataset (Recall@1 = 0.909, Recall@3 = 1.000) and exhibits the strongest cold-start resilience: degradation from 100% to 40% training data is only 34.9%, roughly half that of KGAT (62.8%) and CB-KG (61.7%). At 40% data, PCA-GAT (Recall@1 = 0.591) exceeds KGAT (0.335) by 76.4%. On public benchmarks without constraints, PCA-GAT maintains competitive performance (second-best overall), confirming the constraint mechanism is purely additive. |
| **RQ2** | KG is the foundation: removing it causes −61.04% Recall@5 and −73.04% NDCG@5. Constraints provide additional value (−10.03% Recall@5 when removed). Adaptive gating is |
| | essential: raw injection without gating (Recall@5 = 0.8731) performs worse than no constraints at all (0.8997), while gated injection achieves perfect Recall@5 = 1.000. |
| **RQ3** | All four constraint types individually improve over the no-constraint baseline (+3.84% to +8.68% on Recall@1). Material-operation constraints receive the highest learned weight ( $\lambda$=0.3876 , ~2.8× others). The optimal initial constraint strength is $\lambda_{init}$=0.3, achieving both the best accuracy and lowest variance (std = 0.0064). |
| **RQ4** | d = 64 and L = 1 are optimal. Performance peaks at d = 64 and degrades at d = 128 due to overfitting on the small-scale dataset. Single-head attention (H1) consistently outperforms two-head (H2), as PCA-GAT's constraint-aware mechanism already captures diverse relational semantics within a single head. Deeper propagation (L > 1) hurts due to over-smoothing. CSR = 1.0 across all configurations. |
| **RQ5** | 100% of recommendations have traceable CKG explanation paths. The model exhibits adaptive reasoning: feature-diverse paths for multi-feature parts (e.g., transmission shafts) and plan-diverse paths for single-dominant-feature parts (e.g., spur gears). Learned $\lambda$ weights form a meaningful hierarchy consistent with domain expertise ( $\lambda_{mat} \gg \lambda_{prec} \approx \lambda_{seq} \approx \lambda_{feat}$ ).Constrained edges receive 78% higher attention, with 66.67% path activation rate. |

# 5.Discussion

The experimental results in Section 4 provide comprehensive evidence for the effectiveness of PCA-GAT. In this section, we move beyond performance numbers to discuss the broader implications of our findings (Section 5.1), justify key design choices against alternative approaches (Section 5.2), examine industrial deployment feasibility (Section 5.3), and identify limitations with corresponding future directions (Section 5.4).

## 5.1 Key Findings

Four implications emerge from the experimental evidence presented in Section 4, each extending beyond the specific performance gains to inform broader research directions.

**Implication 1: In manufacturing recommendation, the bottleneck is knowledge representation, not algorithmic capacity.** The complete failure of pure CF methods despite an interaction density 16× higher than Amazon-Book (Section 4.2.1, Finding 4) reveals that the limiting factor is absolute interaction

volume, not algorithmic sophistication. This finding carries a practical message for manufacturing enterprises: investing in richer knowledge representation (i.e., constructing and maintaining high-quality KGs) is likely to yield greater returns than adopting more complex recommendation algorithms. More broadly, the CF framework's value in manufacturing lies in providing a principled ranking-learning objective and standardized evaluation protocol, not in collaborative signals per se.

**Implication 2: Factual knowledge and normative knowledge require independent processing channels.** The ablation results (Section 4.3, Finding 2) demonstrate that normative rules ("45 steel SHOULD-USE turning") encode information absent from factual KG topology ("45 steel IS-A carbon steel"). This distinction is not manufacturing-specific — it applies to any domain where decisions follow objective rules rather than subjective preferences, such as clinical treatment planning (drug-condition compatibility), legal compliance checking (regulation-action constraints), or engineering design (material-load constraints). The general principle is that when a KG contains both descriptive and prescriptive knowledge, collapsing them into a single representation limits the model's discriminative capacity.

**Implication 3: Domain constraints function as data-independent inductive priors whose value scales inversely with data availability.** The monotonic widening of PCA-GAT's advantage over baselines as training data decreases (Section 4.2.1, Finding 2) reveals a complementary relationship between data-driven learning and rule-based knowledge. This property has direct practical relevance: in manufacturing, process rules have been accumulated over decades, but adoption data for new part families is inherently limited. More generally, this finding suggests that for any cold-start scenario where domain rules exist, injecting them as learnable attention biases — rather than hard filters or auxiliary losses — offers a principled way to bridge the data gap while preserving the model's ability to refine rule application through learning.

**Implication 4: Standardized evaluation is itself a methodological contribution.** The cross-category benchmarking in Section 4.2 revealed insights invisible to prior assessment approaches: the complete failure of pure CF despite high interaction density, the KGIN paradox (leading on public benchmarks yet failing in manufacturing), and the convergent degradation of architecturally distinct methods (CB-KG and KGAT). None of these findings could have been obtained through the case-study or within-category evaluation approaches prevalent in prior manufacturing process recommendation research (Table 1). This underscores that establishing a shared evaluation protocol — not just proposing a new model — is essential for methodological progress in domain-specific recommendation.

Taken together, these implications point to a paradigm shift in manufacturing process recommendation — from "engineers seeking knowledge" to "knowledge reaching engineers." Under the conventional retrieval-and-matching paradigm, process engineers must formulate queries and evaluate candidate plans individually; PCA-GAT, by contrast, automatically generates a ranked recommendation list from a part's KG-encoded attributes and, through the constraint-driven explanation paths discussed in Section 4.6, repositions the engineer's role from plan searcher to plan reviewer.

## 5.2 Design Justification

A natural question is whether simpler approaches could achieve comparable results. We address this by examining three alternative constraint integration strategies against our experimental evidence.

**Why not post-processing filtering?** Post-processing generates recommendations from a constraint-unaware model, then removes constraint-violating plans. The fundamental limitation is that constraints cannot influence representation learning: constraint-violating plans may occupy embedding positions near target parts, and while filtering can remove them from the final list, it cannot correct the distorted embedding geometry. The w/o Constraint ablation (Table 10) quantifies this cost: without constraint guidance during training, Recall@1 drops by 13.06% and NDCG@5 by 13.22%. In sparse manufacturing scenarios, this distortion is amplified because limited interaction data provides insufficient signal to push incompatible entities apart.

**Why not encode constraints as KG edges?** This graph augmentation paradigm (Section 2.3) adds constraints as additional CKG edges. Two issues arise. First, our ~1,200 augmented constraint entries constitute a small fraction of total CKG edges (>10,000), and their influence is progressively diluted through multi-layer propagation—consistent with the performance degradation observed when increasing L from 1 to 4 (Table 14). Second, binary edges cannot capture the continuous strength and polarity that distinguish “strongly suitable” (+0.8) from “marginally suitable” (+0.2). The constraint strength experiment (Table 12) confirms that this continuous representation matters: performance varies substantially across λ_init values, with a clear optimum at 0.3.

**Why is adaptive gating necessary?** The w/o Gate ablation (Table 10) provides the most compelling evidence: raw constraint injection ($g_{ij}$ = 1) achieves Recall@5 = 0.8731, worse than having no constraints at all (0.8997). This counterintuitive result reveals a fundamental tension: constraints are general rules, but manufacturing decisions are context-dependent. The constraint “high-precision surfaces require grinding” is correct as a general principle, but blindly enforcing it suppresses valid alternatives such as hard turning for certain hardened steels. The adaptive gate resolves this by learning to open (≈1.0) when a constraint is highly relevant and close (≈0.0) when it may be misleading. This finding carries a broader methodological message: in any domain where expert knowledge is integrated into learning models, context-adaptive modulation is likely essential—rigid injection risks being counterproductive.

**Generalizability to constraint-free scenarios.** Results on three public benchmarks (Table 8) validate the self-regulating property of the adaptive gating mechanism: when constraint scores are uniformly zero, gate activations converge to near-zero values, and the framework automatically reverts to purely data-driven attention propagation. This confirms that the constraint-aware architecture introduces no representational overhead or optimization interference in the absence of domain knowledge—a necessary condition for cross-domain applicability. Notably, KGIN leads on public datasets through intent decomposition but fails dramatically in manufacturing (Recall@1 = 0.504 vs. PCA-GAT's 0.909), empirically confirming that deterministic engineering rules require constraint injection rather than intent modeling.

## 5.3 System Integration Architecture for Industrial Deployment

To bridge the gap between algorithmic innovation and manufacturing practice, we developed a prototype system that embeds PCA-GAT into existing enterprise information infrastructure. As illustrated in Fig. 10, the architecture adopts a three-layer design aligned with the ISA-95 standard, targeting deployment within Manufacturing Execution Systems (MES) and Product Data Management (PDM) environments.

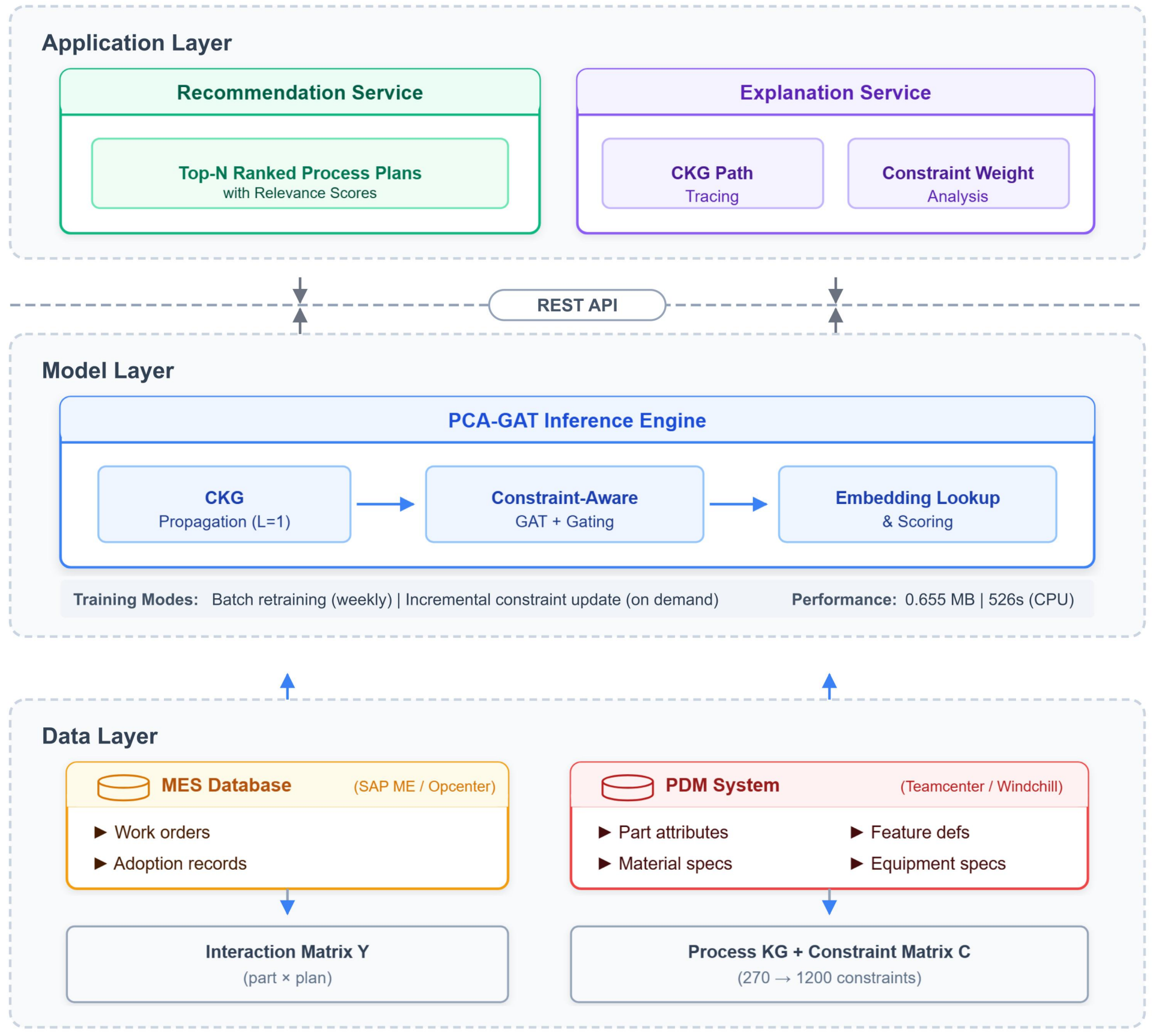

Fig. 10. System integration architecture for deploying PCA-GAT in manufacturing enterprises.

The **Data Layer** interfaces with two standard enterprise data sources: an interaction data pipeline that extracts historical part-plan adoption records from MES work orders, requiring no additional data collection; and a knowledge graph pipeline that populates the process KG from structured PDM data — including part-material associations, feature definitions, and equipment specifications — mapped to KG triplets via schema alignment. Constraint entries are currently extracted through expert annotation (Section 3.3.1), though LLM-assisted extraction represents a promising path to reduce this effort (Section 5.4).

The **Model Layer** hosts the PCA-GAT inference engine. The compact model footprint and fast training time reported in Section 4.5.1 confirm that the model can run on standard server hardware without dedicated GPU infrastructure. Two operational modes have been validated: (i) batch retraining, triggered when new part-plan interactions accumulate; and (ii) incremental constraint update, which modifies only the constraint matrix and retrains constraint-related parameters ($\lambda_c$, $\lambda_g$, gate MLP) while freezing the main GNN parameters. Both modes complete within minutes on commodity hardware.

The **Application Layer** provides two services: a Recommendation Service that accepts a part's CKG-encoded attributes and returns a ranked list of Top-N candidate process plans with relevance scores, and an Explanation Service that provides path-level interpretability (Section 4.6) and constraint-level weight analysis (Section 4.6.4).

The prototype has been developed and validated using real-world data from the collaborating aerospace manufacturing enterprise. All data pipelines have been implemented and tested with production-format data exports, and the recommendation and explanation services have been encapsulated as RESTful APIs with verified compatibility with standard MES/PDM data schemas. Full production deployment, including user acceptance testing with process engineers, constitutes the immediate next step.

## 5.4 Limitations and Future Work

We acknowledge four limitations that define directions for future research.

**Dataset scale and diversity.** The machining dataset contains 115 parts and 507 plans from a single aerospace enterprise (Table 3). While this scale is representative of real manufacturing scenarios where part families are inherently limited, it constrains the statistical power of our conclusions. In particular, the perfect Recall@5 and NDCG@5 achieved under full training data suggest that the candidate space may be insufficiently large to stress-test ranking quality at deeper positions. Whether the learned $\lambda$ hierarchy ($\lambda_{mat} \gg \lambda_{prec} \approx \lambda_{seq} \approx \lambda_{feat}$; Section 4.6.4) generalizes to other manufacturing sectors — where different constraint types may dominate — remains an open empirical question. Validation on datasets from diverse manufacturing domains (e.g., sheet metal forming, additive manufacturing, electronics assembly) is needed. We note that the constraint annotation protocol and PCA-GAT architecture are domain-agnostic by design, requiring only the definition of relevant constraint types and expert-assigned strength values for a new domain.

**Manual constraint annotation.** The 270 raw constraint entries were annotated by two senior engineers through a two-round protocol. While quality is ensured (ICC = 0.87 before reconciliation; Section 3.3.1), this process is difficult to scale. Future work could leverage LLMs to extract constraint triplets from unstructured sources such as process handbooks and machining standards, with expert review serving as quality assurance rather than primary authoring. Recent progress in LLM-driven KG construction [32,14] suggests that such automation is increasingly feasible. Our finding that the learnable $\lambda$ mechanism automatically discovers constraint importance (Section 4.4.1, 4.6.4) suggests that even imperfect, automatically extracted constraints may provide substantial value if the model can learn to weight them appropriately.

**Static constraint assumption.** The current framework treats constraints as fixed inputs determined before training. In practice, manufacturing constraints evolve as new materials, equipment, and standards emerge. Developing an incremental update mechanism — leveraging PCA-GAT's modular architecture where constraint parameters ($\lambda_c$, $\lambda_g$, gate MLP) are decoupled from main GNN parameters — is a natural extension. The compact model footprint (Section 4.5.1) makes frequent retraining computationally feasible even in resource-constrained environments.

**Architectural scope.** The current implementation instantiates the constraint-aware attention paradigm using a GAT-based propagation backbone. Whether this paradigm transfers effectively to other GNN architectures — for instance, combining constraint injection with intent decomposition [28] or contrastive learning frameworks [36,37] — remains to be validated. The core principle of maintaining a separate channel for normative knowledge should be architecture-agnostic, but empirical confirmation across backbone architectures is needed.

**Cross-domain empirical validation.** While we have discussed the potential applicability of constraint-aware attention to multiple engineering domains (Section 5.1), empirical validation in this study is limited to the manufacturing scenario. Future work should instantiate the framework in at least one additional engineering domain — such as clinical treatment recommendation or building code compliance checking — to provide direct empirical evidence for cross-domain generalizability. The framework's modular design, where constraint types and strength values are decoupled from the model architecture, is specifically intended to facilitate such cross-domain instantiation.

# 6.Conclusions

This paper addresses the challenge of integrating factual and normative knowledge in knowledge-graph-based engineering decision support. Using machining process plan recommendation as an exemplar task, we formulate the problem as a KG-enhanced collaborative filtering problem and propose a Process Constraint-Aware Graph Attention Network (PCA-GAT), in which four types of domain constraints — material–operation compatibility, precision–operation requirements, feature–operation mapping, and operation sequencing — are injected as continuous-valued attention biases during graph propagation, with type-specific learnable weights and an adaptive gating mechanism. We further establish the first standardized recommendation evaluation protocol in the engineering process recommendation domain, benchmarking seven methods spanning three categories.

The principal findings are as follows:

(1) Factual knowledge and normative knowledge require independent processing channels. Process constraints provide a 13.06% Recall@1 improvement over KG topology alone, demonstrating that normative rules ("should use") encode information absent from factual relations ("is related to"). Critically, adaptive gating is essential: raw constraint injection without gating degrades performance below the constraint-free baseline, indicating that context-dependent modulation is a prerequisite for beneficial constraint integration. Moreover, on three public benchmarks where no domain constraints exist, the adaptive gating mechanism autonomously deactivates, and PCA-GAT retains fully competitive accuracy, confirming that the constraint-aware components are purely additive with zero degradation in constraint-absent scenarios.

(2) Knowledge graph enrichment is the foundation for overcoming interaction sparsity in engineering decision support. Removing the KG causes a 61.04% drop in Recall@5, confirming that with only 371 training interactions, pure collaborative filtering lacks sufficient signal. Meanwhile, pure CF methods (BPR-MF, LightGCN, CKE) fail entirely in this regime despite interaction density 16× higher than consumer benchmarks — revealing that the primary bottleneck is knowledge representation rather than algorithmic capacity, a result that could not have been established without the cross-category evaluation protocol introduced in this work.

(3) Domain constraints function as data-independent inductive priors whose value scales inversely with data availability. PCA-GAT's Recall@1 degradation from 100% to 40% training data is only 34.9%, roughly half that of the strongest baseline KGAT (62.8%). This cold-start resilience directly addresses the practical scenario where decision rules have been accumulated over decades but historical records for new entities are inherently scarce.

(4) The learned constraint weights form a meaningful hierarchy consistent with domain expertise, with material–operation compatibility receiving the highest weight ($\lambda = 0.3876$, nearly three times that of other types). This confirms that the learnable mechanism captures genuine domain knowledge through end-to-end training without explicit supervision on constraint importance, providing constraint-driven explainable recommendations.

These findings carry implications beyond manufacturing. The core principle — that descriptive and prescriptive knowledge require independent processing channels, and that context-adaptive gating is essential for beneficial constraint integration — applies to any engineering domain where decisions follow codified rules with continuous strengths, such as clinical treatment planning, structural engineering design, or regulatory compliance checking. The constraint-aware architecture's zero-degradation property in constraint-absent scenarios ensures that it can serve as a general-purpose engineering knowledge integration paradigm without domain-specific architectural modification.

From a deployment perspective, the model requires only 0.655 MB memory and 526 seconds of training time, supporting edge deployment and frequent retraining in resource-constrained environments. A prototype system following the ISA-95 three-layer architecture has been developed and validated with real-world enterprise data.

Several limitations point to future directions. The dataset originates from a single aerospace enterprise; whether the learned constraint hierarchy generalizes to other manufacturing sectors remains an open question. The 270 raw constraint entries rely on manual annotation by domain experts; LLM-assisted constraint extraction represents a promising path to improve scalability. The current framework treats constraints as static inputs; developing incremental update mechanisms that leverage PCA-GAT's modular architecture is a natural extension. Validating the constraint-aware attention paradigm on alternative GNN backbones would further establish its generality. Finally, while we have discussed the applicability of constraint-aware attention to multiple engineering domains, empirical validation in this study is limited to the manufacturing scenario; instantiating the framework in additional domains such as clinical decision support would provide direct evidence for cross-domain generalizability.

# Data availability statement

The three public benchmark datasets (Amazon-Book, Yelp2018, Last-fm) used in this study are openly available at [https://github.com/xiangwang1223/knowledge_graph_attention_network]. The machining process dataset is subject to a non-disclosure agreement (NDA) and thus not available for public release. Access may be granted only after approval from the collaborating enterprise via contacting the corresponding author. The de-identified source code, including model implementations, training pipelines, and evaluation scripts, is publicly available at [https://github.com/aeon666-cyt/PCA-GAT/tree/main] upon publication.

## Declaration of competing interest

The authors declare that they have no known competing financial interests or personal relationships that could have appeared to influence the work reported in this paper.

## Declaration of generative AI and AI-assisted technologies in the manuscript preparation process

During the preparation of this work the author used Gemini (Gemini 3.1 Pro) in order to improve the readability and language quality of the manuscript. After using this tool, the author reviewed and edited the content as needed and takes full responsibility for the content of the published article.